\documentclass[10pt,twocolumn,letterpaper]{article}

\usepackage{cvpr}              % To produce the CAMERA-READY version

\usepackage{overpic}
\usepackage[dvipsnames]{xcolor}

\definecolor{cvprblue}{rgb}{0.21,0.49,0.74}
\usepackage[pagebackref,breaklinks,colorlinks,citecolor=cvprblue]{hyperref}

\def\paperID{7876} % *** Enter the Paper ID here
\def\confName{CVPR}
\def\confYear{2024}

\title{FlashAvatar: High-fidelity Head Avatar with Efficient Gaussian Embedding}
\usepackage[accsupp]{axessibility}

\author{Jun Xiang \qquad Xuan Gao \qquad Yudong Guo \qquad Juyong Zhang\thanks{Corresponding author}\\
University of Science and Technology of China\\
{\tt\small \{junxiang@mail., gx2017@mail., yudong@, juyong@\}ustc.edu.cn}
}

\begin{document}
% \maketitle
% \author{
%     Jun Xiang\textsuperscript{1} \quad
%     Xuan Gao\textsuperscript{1} \quad
%     Yudong Guo\textsuperscript{2} \quad
%     Juyong Zhang\textsuperscript{1} \\[5pt]
%     $^1$University of Science and Technology of China \qquad
%     $^2$Image Derivative Inc  \\[8pt]
% }

% \author{
%     Jun Xiang \quad
%     Xuan Gao \quad
%     Yudong Guo \quad
%     Juyong Zhang \\[5pt]
%     University of Science and Technology of China \\[8pt]
% }

% teaser
\twocolumn[{
\maketitle
\vspace*{-9mm}
\begin{center}
   \begin{overpic}
        [width=\linewidth]{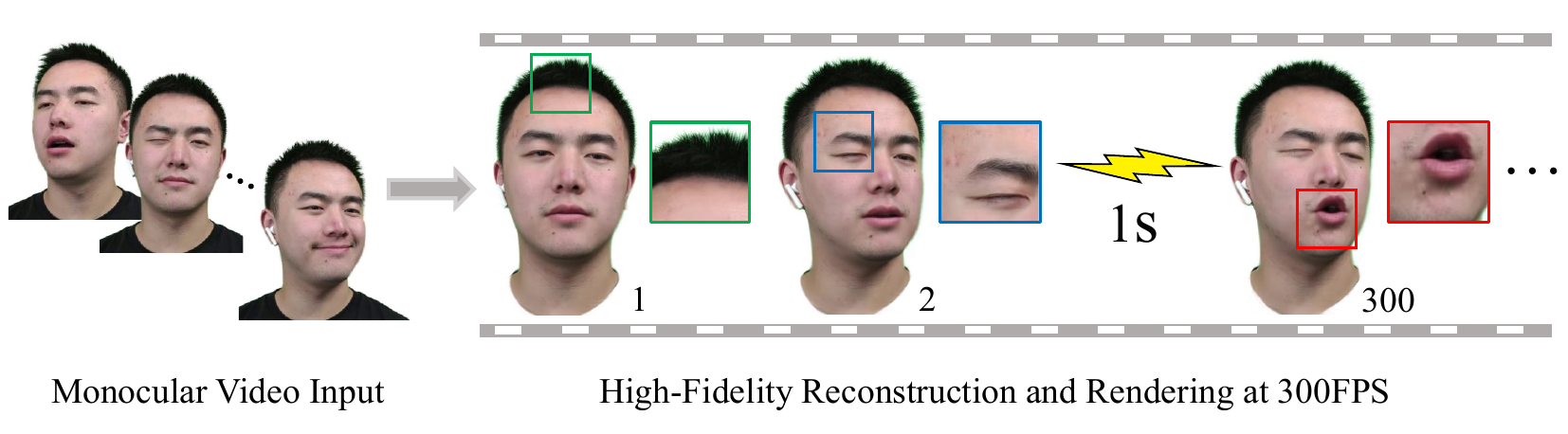}
   \end{overpic}
\end{center}
\vspace*{-5mm}
\captionof{figure}{Given a monocular video sequence, our proposed FlashAvatar can reconstruct a high-fidelity digital avatar in minutes which can be animated and rendered over 300FPS at the resolution of $512\times 512$ with an Nvidia RTX 3090.}
\label{fig:teaser}
\vspace*{5mm}
}]

{
  \renewcommand{\thefootnote}%
    {\fnsymbol{footnote}}
  \footnotetext[1]{Corresponding Author}
}

\begin{abstract}
We propose FlashAvatar, a novel and lightweight 3D animatable avatar representation that could reconstruct a digital avatar from a short monocular video sequence in minutes and render high-fidelity photo-realistic images at 300FPS on a consumer-grade GPU. To achieve this, we maintain a uniform 3D Gaussian field embedded in the surface of a parametric face model and learn extra spatial offset to model non-surface regions and subtle facial details. While full use of geometric priors can capture high-frequency facial details and preserve exaggerated expressions, proper initialization can help reduce the number of Gaussians, thus enabling super-fast rendering speed. Extensive experimental results demonstrate that FlashAvatar outperforms existing works regarding visual quality and personalized details and is almost an order of magnitude faster in rendering speed. Project page: \href{https://ustc3dv.github.io/FlashAvatar/}{https://ustc3dv.github.io/FlashAvatar/}
\end{abstract}    
\section{Introduction}
\label{sec:intro}
% With the expansion of the Internet and the development of AR/VR technologies, interpersonal communication is gradually shifting from offline face-to-face interactions to online discussions and even interactions in virtual world. To achieve this, 

Achieving low-cost, high-fidelity digital humans with real-time multi-modal interaction, natural expressions and movements, \etc, is a key underlying technology for many AR and VR applications, such as immersive remote conferencing. With this target in mind, this work aims to present a high-fidelity animatable head avatar that enables efficient reconstruction and lightning-fast rendering, such that the remaining computing resources can support other interactive tasks of multi-modal digital humans.

Previous works have made notable progress, while there still exist some shortcomings. 3D morphable models (3DMMs)~\cite{li2017learning,paysan20093d} based methods~\cite{kim2018deep,grassal2022neural,khakhulin2022realistic} are compatible with the standard graphics pipeline and can extrapolate to unseen deformations. However, the limitations of relying on coarse geometry and fixed topology prevent them from modeling complex hairstyles or accessories like eyeglasses. Works~\cite{gafni2021dynamic,zheng2022avatar,hong2022headnerf,athar2022rignerf,bergman2022generative,jiang2022selfrecon,guo2021adnerf} building on neural implicit representations~\cite{mildenhall2020nerf,mescheder2019occupancy,park2019deepsdf} could well capture fine features with great rendering quality and 3D consistency but commonly suffer from slow training and inference computation speed. 
Motivated by works~\cite{sun2022direct,fridovich2022plenoxels,chen2022tensorf,muller2022instant,liu2020neural} for accelerating Neural Radiance Field (NeRF)~\cite{mildenhall2020nerf} rendering, \cite{gao2022reconstructing,zielonka2023instant,xu2023avatarmav} apply voxel representations like voxel grids and multi-level hash tables to speed up head avatar reconstruction. Nevertheless, the volume rendering mechanism of excessive sampling and alpha composition still limits the inference speed.

Recently, 3D Gaussian Splatting (3D-GS)~\cite{kerbl3Dgaussians} revolutionized radiance field rendering by introducing non-neural 3D Gaussians as geometric primitives and developing a fast rendering algorithm that supports anisotropic splatting.
% The non-neural nature of it reminds us that combining it with concrete 3DMM mesh will be a new solution to avatar representation. PointAvatar~\cite{zheng2023pointavatar} follows similar guidance by using point cloud as the primary representation. Nevertheless, 3D Gaussian allows anisotropic splatting and fast back-propagation, which is undoubtedly more expressive and easy to optimize.
Follow-up works~\cite{yang2023deformable3dgs,wu20234dgaussians} of 3D-GS have already extended it to dynamic scenes by maintaining a canonical Gaussian field and constructing another deformation field conditional on timestamp. However, our experiments have demonstrated that this ``canonical + deformation" strategy cannot robustly model dynamic head avatar with complex expressions even if we replace the condition with more meaningful expression code.
% considering the intricacies of facial movement.
% But considering the intricacies of facial movement, how to model dynamic head avatar precisely and efficiently with 3D-GS is still an unexplored problem. 

Based on these observations, we propose a novel avatar representation named \emph{FlashAvatar}. We initialize a mesh-embedded Gaussian field to model the avatar's main appearance and facial expressions and learn extra offset to model non-surface features and small facial dynamics. Specifically, we initially attach 3D Gaussians to the mesh surface, which will move along with the mesh. In this way, we do not need to learn large deformations caused by expression changes.
% , which is the main learning burden of existing methods~\cite{gafni2021dynamic,zheng2023pointavatar}. 
However, coarse mesh geometry does not involve non-surface regions like hair or fine facial details like wrinkles. Thus, we use an additional offset network to predict the spatial offsets of 3D Gaussians.

\begin{figure}
  \centering
  \includegraphics[width=0.92\linewidth]{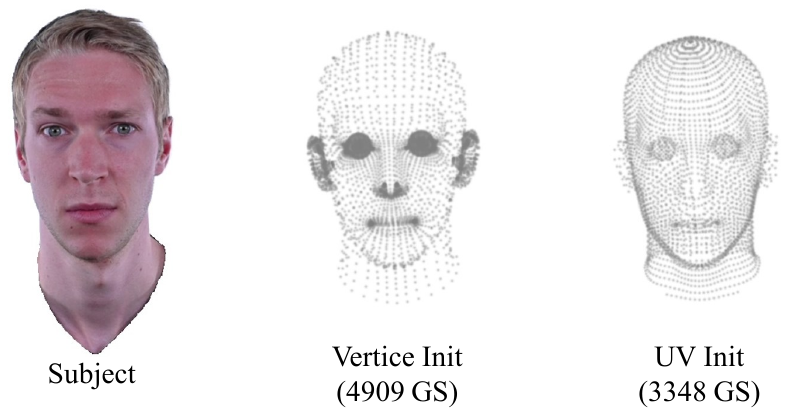}
  \caption{Initialization in UV space corresponds to a more uniform Gaussian position distribution, which could model full head details better. We only sample points in the head region, including neck, so the number of sample vertices is smaller than FLAME vertice number 5023.}
  \label{fig:init}
\end{figure}

% To determine the initial position of Gaussians on mesh surface, we conduct a flexible UV sampling. Compared with direct sampling on faces or Binding Gaussians to vertices, UV sampling is more uniform and convenient, which supports flexible density control of sampling and leads to better results.
While attaching Gaussians to 3D mesh vertices is a quite straightforward strategy, it is hard to recover complete surface information since the position distribution of vertices is highly uneven. Direct sampling on mesh faces has the same problem of unevenness.
Instead, we conduct a flexible UV sampling and turn to maintain a canonical Gaussian field in the UV space. This sampling strategy supports easy density control of Gaussians and generates a much more uniform position distribution (see~\cref{fig:init}), which leads to better reconstruction results.

With the help of uniform UV sampling and critical mesh-attached initialization, we achieve photo-realistic head avatar representation with as few 3D Gaussians as possible. Compared with existing 3DMM-based methods, mesh topology will not restrict our representation as tracked meshes only provide initial position distribution and serve as motion-driven tools. Compared to works building on neural implicit representation, we fully introduce geometric priors, exploit the potential of Gaussian-based radiance field, and thus enable super-fast training and inference. In summary, our contributions include the following aspects:
\begin{itemize}
\item We combine Gaussian splats with 3D parametric face model by attaching the Gaussians to the mesh surface and learning extra offsets to model detailed facial dynamics and non-facial features, which leverages dynamic and geometric priors to a great extent and increases the training efficiency.
\item Our uniform and flexible UV sampling enables optimal mesh-based initialization, which compresses Gaussian number to 10K level and helps achieve a stable rendering speed at 300FPS at the resolution of $512\times512$. 
% while ensuring the SOTA rendering quality.
\item Experiments demonstrate the high fidelity of our approach even on challenging cases, recovering almost all fine facial details, thin structures, and subtle expressions.
\end{itemize}

\section{Related Work}
\label{sec:related}
%-------------------------------------------------------------------------
\subsection{Digital Head Model}

Digital head model could be classified into explicit and implicit representations. Explicit representations based on mesh have a long history of development. 3DMM \cite{blanz1999morphable} first embeds 3D head shape into several low-dimensional PCA spaces. After that, many works~\cite{cao2013facewarehouse,vlasic2006face,yang2020facescape,ranjan2018generating,tran2018nonlinear,guo20213d,li2017learning,zhang2023hack} are proposed and used for improvement of representation ability.
Recently, \cite{kim2018deep,thies2019deferred,thies2020neural} adopt 2D neural rendering for photo-realistic portrait synthesis but either ignore non-facial regions or suffer from temporal and spatial inconsistencies due to their loose bound to the 3D geometry.
\cite{grassal2022neural,khakhulin2022realistic,DECA:Siggraph2021} opt to learn vertex offset on the head geometry to reconstruct the detailed head model.
% But because of the limited representation ability of mesh model and approximated differentiable rendering, both geometry and texture artifacts may occur in hair, eyes, and mouth regions.
However, geometry and texture artifacts may occur in hair, eyes, and mouth regions because of the limited representation ability of the mesh model and the approximated differentiable rendering.
PointAvatar~\cite{zheng2023pointavatar} proposes a deformable point-based representation, which breaks through the limitation of mesh-based models but needs excessive points and long-time training.
Implicit head models use neural functions to represent digital head avatars. There have been extensive works on personalized head modeling~\cite{athar2022rignerf,gafni2021dynamic,zheng2022imface,zheng2022avatar}. They tend to maintain high fidelity but must be more efficient in training or inference.
% Local feature grid is found to be helpful to reduce the learning burden of MLP and accelerate the rendering~\cite{Gao2022nerfblendshape,xu2023avatarmav,zielonka2022instant}. There are also works using volumetric primitives~\cite{Lombardi21} and neural rendering~\cite{hong2022headnerf} to improve inference efficiency.
\cite{Lombardi21} uses volumetric primitives to improve inference efficiency, and \cite{gao2022reconstructing,xu2023avatarmav,zielonka2023instant} use local feature grid to reduce the learning burden of neural network and accelerate the training process.
To our knowledge, our work is the first to introduce a mesh-guided Gaussian field for modeling head avatars.

% Most 3D avatars reconstruction methods either build on 3DMMs or apply neural implicit representations.
% Works~\cite{kim2018deep,grassal2022neural,khakhulin2022realistic} building on 3DMMs jointly optimize texture and geometry in an analysis-by-synthesis fashion with the help of standard computer graphics pipelines for rasterizing meshes.
% For example, NHA~\cite{grassal2022neural} predicts vertex offsets of FLAME mesh as well as a view- and expression-dependent texture.
% Methods~\cite{gafni2021dynamic,zheng2022avatar,athar2022rignerf} based on neural implicit representations, especially NeRF~\cite{mildenhall2020nerf}, always outperform mesh-based methods in modeling non-surface parts like earrings and hair strands.
% \cite{gao2022reconstructing,zielonka2023instant,xu2023avatarmav} make further training acceleration by utilizing voxel representation. (Among them, INSTA...).
% PointAvatar~\cite{zheng2023pointavatar} takes a different path by using point cloud as basic representation which needs excessive points and long-time training. To the best of our knowledge, our work is the first to introduce mesh-based Gaussian field for modeling head avatars.

%-------------------------------------------------------------------------
\subsection{Scene representations with 3D-GS}

3D Gaussian Splatting~\cite{kerbl3Dgaussians} is currently the SOTA method of scene reconstruction and novel view synthesis regarding rendering speed and visual quality, which inspires a series of works.
% DreamGaussian~\cite{tang2023dreamgaussian} and GaussianDreamer~\cite{GaussianDreamer} adapt 3D-GS into 3D generative tasks by optimizing 3D gaussians using score distillation sampling (SDS)~\cite{poole2022dreamfusion}.
\cite{tang2023dreamgaussian,GaussianDreamer,chen2023text} adapt 3D-GS into 3D generative tasks by optimizing Gaussian field using score distillation sampling (SDS)~\cite{poole2022dreamfusion}.
DreamGaussian~\cite{tang2023dreamgaussian} also designs an efficient mesh extraction algorithm for the Gaussian field. Dynamic3DGS~\cite{luiten2023dynamic} first extends 3D-GS to model dynamic scenes, reconstructing the ``point cloud" frame by frame. Different from~\cite{luiten2023dynamic}, Deformable3DGS~\cite{yang2023deformable3dgs} and 4D-GS~\cite{wu20234dgaussians} focus on monocular dynamic scene reconstruction. They both maintain a canonical 3D Gaussian space and optimize an additional deformation field conditional on timestamp. Our work uses 3D-GS to represent dynamic head avatars with complex facial alterations. Rather than adopting the ``canonical + deformation'' strategy, we attach 3D Gaussians to the head mesh and learn dynamic offsets to model photo-realistic avatars.

%-------------------------------------------------------------------------
\subsection{Radiance field acceleration}
Neural radiance field (NeRF)~\cite{mildenhall2020nerf} and follow-up works~\cite{kaizhang2020,wang2021neus,niemeyer2021giraffe,barron2022mip} significantly develop scene representation but suffer from low rendering efficiency. To accelerate radiance field training and rendering, most works make full use of voxel-based structures like octree~\cite{fridovich2022plenoxels,liu2020neural} and voxel grid~\cite{sun2022direct,garbin2021fastnerf,hedman2021baking} by baking information into them which usually needs large cache. INGP~\cite{muller2022instant} adopts a more compressed compact data structure (\ie multi-resolution hash table) and achieves a speedup of several orders of magnitude on training speed but struggles to achieve the visual quality obtained by SOTA NeRF methods~\cite{barron2022mip}. Recently, 3D-GS~\cite{kerbl3Dgaussians} replaces neural primitives with non-neural 3D Gaussians and designs a fast tile-based rasterizer for Gaussian splats, which guarantees both quality and speed. We apply it to dynamic head representation. Via rational position initialization and density control for Gaussians, we significantly compress the number of used Gasussians and achieve instant training and a stable rendering frame rate at 300FPS.
%-------------------------------------------------------------------------

% \section{Preliminary}
\section{Background}
\label{sec:preliminary}
\begin{figure*}[htb]
  \centering
  \includegraphics[width=1\linewidth]{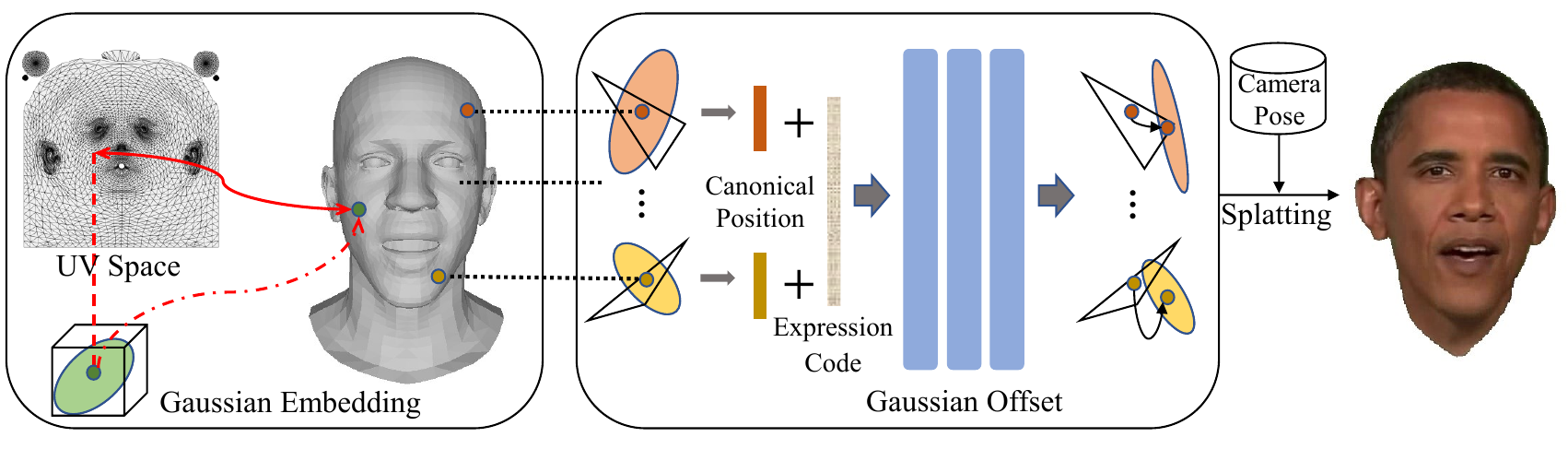}
  \caption{Overview. We initially maintain the 3D Gaussian field in 2D UV space and embed them into dynamic FLAME mesh surfaces through mesh rasterization. For every surface-embedded 3D Gaussian, the offset network takes tracked expression code and the corresponding position of the Gaussian center on canonical mesh as input, outputs the spatial offset, including position, rotation, and scaling deformation. The deformed Gaussians are then splatted to render the image with a given pose.}
  \label{fig:pipeline}
\end{figure*}
%-------------------------------------------------------------------------
\textbf{3D Gaussian Splatting.}
Different from previous methods \cite{kopanas2021point,yifan2019differentiable}, which use 2D points with normals to represent a scene, 3D-GS \cite{kerbl3Dgaussians} chooses 3D Gaussians as geometric primitives of scenes. Every Gaussian is defined by a 3D covariance matrix $\mathbf{\Sigma}$ centered at point $\mathbf{\mu}$:
\begin{equation}
  g(\mathbf{x}) = e^{-\frac{1}{2} (\mathbf{x}-\mathbf{\mu})^{T} \mathbf{\Sigma} ^{-1}(\mathbf{x}-\mathbf{\mu})}
  \label{eq:gs}
\end{equation}
To enable differentiable optimization, the positive semi-definite matrix $\mathbf{\Sigma}$ can be decomposed into a rotation matrix $\mathbf{R}$ and a scaling matrix $\mathbf{S}$ corresponding to learnable quaternion $\mathbf{r}$ and scaling vector $\mathbf{s}$:
\begin{equation}
  \mathbf{\Sigma} = \mathbf{R}\mathbf{S}\mathbf{S}^T\mathbf{R}^T
  \label{eq:cov}
\end{equation}
Given a viewing transformation $W$ and the Jacobian $J$ of the affine approximation of the projective transformation, 3D Gaussians are projected to 2D space for rendering following \cite{zwicker2001ewa}:
\begin{equation}
  \mathbf{\Sigma}^{\prime} = JW\mathbf{\Sigma} W^TJ^T
  \label{eq:project}
\end{equation}
Besides spatial parameters $\mathbf{\mu}$, $\mathbf{r}$ and $\mathbf{s}$, we attach every 3D Gaussian another two attributes: opacity $o$ and spherical harmonic (SH) coefficients $\mathbf{h}$ representing color $\mathbf{c}$. The final color for a given pixel is calculated by sorting and blending the overlapped Gaussians:
\begin{equation}
  \mathbf{C} = \sum_{i\in N} \mathbf{c}_i\alpha_i\prod_{j=1}^{i-1} (1-\alpha_j)  
  \label{eq:blend}
\end{equation}
where $\alpha_i$ represents the density computed by the 2D Gaussian with covariance $\Sigma^\prime$ multiplied by opacity $o$.

\noindent
\textbf{Analysis.}
The non-neural nature of 3D-GS reminds us that combining it with concrete mesh will be a new solution to avatar representation. PointAvatar~\cite{zheng2023pointavatar} follows similar guidance by using point cloud as the basic representation. In comparison, 3D Gaussian allows anisotropic splatting and fast back-propagation, which is undoubtedly more expressive and easy to optimize.

As in NeRF~\cite{mildenhall2020nerf}, sampled points near the surface of objects always play a critical role in volume rendering. We assume that modeling avatars with 3D Gaussians follows the same rule, and the ideal Gaussian distribution would be concentrated on the head surface. Thus, it motivates us to attach Gaussians to FLAME mesh surface initially.

The densification scheme of 3D-GS helps model general scenes but leads to explosion and uncertainty of Gaussian’s number, which takes more memory consumption and slows down rendering speed.
Since the complexity of head avatars is within a specific range, it is reasonable for us to maintain a fixed number of Gaussians for all subjects instead of adopting the rough splitting strategy of 3D-GS.

\section{Methods}
\label{sec:methods}
Given a monocular video consisting of images $I = \{I_i\}$ along with camera intrinsic parameters $\mathbf{K}$, camera poses $\mathbf{P} = \{P_i\}$ and tracked FLAME~\cite{li2017learning} meshes $\mathbf{M} = \{M_i\}$ with corresponding
% shared shape code $\beta$
expression codes $\Psi = \{\psi_i\}$, we aim to recover high-fidelity head avatars efficiently with great rendering speed.
% We successfully make it by fully utilizing the geometric prior knowledge learned in the face-tracking process and digging into the potential of 3D-GS to represent complex details. By combining two explicit models, we achieve instant training, photo-realistic visual quality and rendering speed over 300 FPS.
By fully utilizing the geometric prior knowledge learned in the face-tracking process and the strong representation ability of 3D-GS, we achieve instant training, photo-realistic visual quality, and rendering speed at 300 FPS. An overview of the proposed model is shown in~\cref{fig:pipeline}.

%-------------------------------------------------------------------------
\subsection{Surface-embedded Gaussian Initialization}

Previous head representations based on implicit functions usually build connections with 3DMM by simply utilizing expression code~\cite{gafni2021dynamic} or transformation of the closest point on mesh between canonical and deformed space~\cite{zielonka2023instant,athar2022rignerf}. In this way, they fail to fully use the geometric priors of mesh. Our solution is to initially attach 3D Gaussians to the mesh surface, which will move along with the mesh, and we conduct this through UV sampling.
% Instead, we choose to sample 3D Gaussians on mesh faces and conduct this by UV sampling.
% It is worth noting that we add additional faces to close the mouth cavity since original FLAME mesh does not model interior mouth.
\begin{figure}
  \centering
  \includegraphics[width=0.9\linewidth]{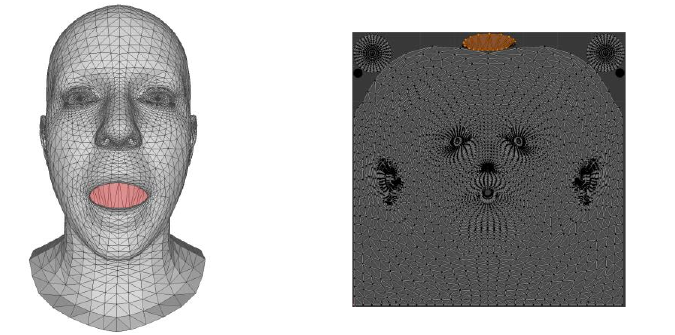}
  \caption{To well model interior mouth, we close the mouth cavity of FLAME mesh with additional faces and broaden up corresponding area on UV map.}
  \label{fig:mouth}
\end{figure}
\begin{figure*}
  \centering
  \includegraphics[width=1.0\linewidth]{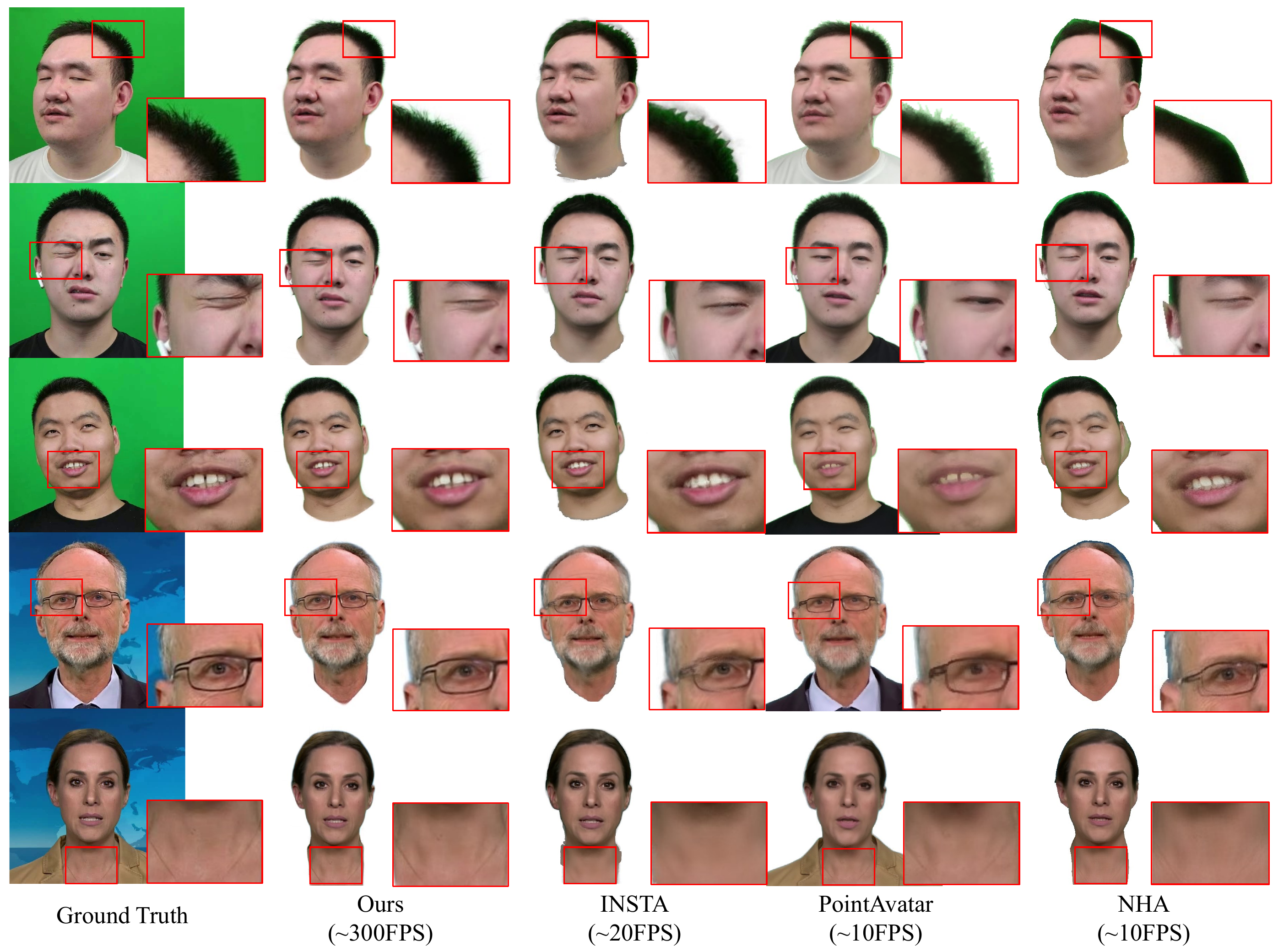}
  \caption{Qualitative comparisons with state-of-the-art head avatar reconstruction methods. Our model well reconstructs facial details, thin structures, and subtle expressions while achieving a remarkable rendering speed over 300FPS.}
  \label{fig:comparison}
\end{figure*}
\noindent

\textbf{UV Sampling.}
We conduct UV sampling to locate Gaussian's position on the mesh surface.
% In other words, We maintain a canonical 3D Gaussian field in 2D UV space.
By rasterizing the FLAME mesh in world space to UV space, we can get a one-to-one correspondence between UV pixels and mesh surface positions.
We sample on the UV map and thus maintain a canonical uniform 3D Gaussian field in 2D UV space.
Since the same mesh topology shares fixed UV parameterization, we only need to conduct rasterization~\cite{ravi2020accelerating} once.
When expression changes, the corresponding 3D position of Gaussians can be obtained by weighting vertex coordinates using fixed barycentric coordinates.
% UV sampling is equal to sampling on mesh faces but more uniform and flexible.

% While attaching Gaussians to 3D mesh vertices is a more straightforward choice, it behaves badly because vertices are located in a raised position with uneven distribution and thus is hard to recover full surface information (see~\cref{fig:init}). In contrast, our initialization strategy generates a much more uniform distribution and has better performance.

We can conveniently control Gaussian density by adjusting UV map resolution, sampling interval, and even the covering area of different parts on the UV map based on semantic correspondence.
For example, we broaden up the interior mouth area on UV considering the complexity of the internal structure of mouth. It is worth noting that we add additional faces to close the mouth cavity since original FLAME mesh does not model interior mouth (see~\cref{fig:mouth}).

According to~\cref{sec:preliminary}, Gaussian field can be parameterized as $G = \{\mathbf{\mu}, \mathbf{r}, \mathbf{s}, o, \mathbf{h}\}$. Through UV sampling, we have defined the initial position of mesh-attached Gaussians $\mathbf{\mu}_{M}$. And in our settings, opacity $o$, SH coefficients $\mathbf{h}$, rotation $\mathbf{r}$ and scaling $\mathbf{s}$ are learnable parameters. While the former two attributes, which decide the main appearance of avatars, converge to be fixed, the last two spatial parameters together with $\mathbf{\mu}_{M}$ are added with extra deformation to model non-surface features as well as dynamic details of the face.
%-------------------------------------------------------------------------
\subsection{Gaussian Offset}

We denote the centers of mesh-attached Gaussians as $\mathbf{\mu}_{M}$ and corresponding positions on canonical mesh $\mathbf{\mu}_T$.
% Since large position deformation caused by expression changes has been modeled by $\mathbf{\mu}_{M}$ compared to $\mathbf{\mu}_T$, we just need to further predict offsets of position $\mathbf{\mu}_{M}$, rotation $\mathbf{r}$ and scaling $\mathbf{s}$.
Even though main position deformation caused by expression changes has been modeled by $\mathbf{\mu}_{M}$ compared to $\mathbf{\mu}_T$, non-surface regions and subtle facial details are not considered, and we model them through further adding dynamic spatial offset to Gaussians.
The offset network is an MLP $F_\theta$ that takes $\mathbf{\mu}_T$ and $\psi$ as input, and outputs spatial residuals of Gaussians:
\begin{equation}
  \Delta\mathbf{\mu}_\psi, \Delta\mathbf{r}_\psi, \Delta\mathbf{s}_\psi = F_\theta (\gamma(\mathbf{\mu}_T), \psi)
  \label{eq:mlp}
\end{equation}
where $\gamma$ denotes the positional encoding as introduced by Mildenhall \etal~\cite{mildenhall2020nerf}.
Then, the final spatial parameters of Gaussians can be computed as:
\begin{equation}
  \mathbf{\mu}_\psi, \mathbf{r}_\psi, \mathbf{s}_\psi = (\mathbf{\mu}_{M} \oplus \Delta\mathbf{\mu}_\psi, \mathbf{r} \oplus \Delta\mathbf{r}_\psi, \mathbf{s} \oplus \Delta\mathbf{s}_\psi)
  \label{eq:offset}
\end{equation}
% While \cite{luiten2023dynamic} argues we can model scene dynamics by merely optimizing the position and rotation parameters, we find changeable scaling parameters make better performance, especially in modeling teeth. This is explainable 
As we initially attach 3D Gaussians to mesh faces, the region a group of Gaussians could influence may expand or shrink with the altering size of mesh faces, especially in the early training process. By adjusting scaling dynamically together with position and rotation, we can better model fixed-size parts like teeth.
%-------------------------------------------------------------------------
\subsection{Training Scheme}
Corresponding to expression $\psi$, our 3D Gaussians field will be $G = \{\mathbf{\mu}_\psi, \mathbf{r}_\psi, \mathbf{s}_\psi, o, \mathbf{h}\}$. And following Equation \cref{eq:blend}, we will get the rendering image $\hat{I}$.

To measure the photometric error, we use Huber loss \cite{huber1992robust} with $\delta = 0.1$:
\begin{equation}
  \mathcal{L}_{H} (x, \hat{x}) = 
  \left\{
  \begin{matrix}
      \frac{1}{2} (x-\hat{x})^2 & \text{if} \; |x-\hat{x}|<\delta \\
      \delta ((x-\hat{x})-\frac{1}{2}\delta) & \text{otherwise}
  \end{matrix}
  \right.
  \label{eq:huber-loss}
\end{equation}
Specifically, we conduct bigger weight for mouth region with mask $\mathcal{M}$, so the photometric loss $\mathcal{L}_C$ is defined as:
\begin{equation}
  \mathcal{L}_{C} = \mathcal{L}_{H}(I, \hat{I}) + \lambda_{\textrm{mouth}}\mathcal{L}_{H}(I\cdot\mathcal{M}, \hat{I}\cdot\mathcal{M})
  \label{eq:color-loss}
\end{equation}
In addition to photometric loss $\mathcal{L}_{C}$, we adopt perceptual loss
$\mathcal{L}_{\textrm{lpips}}$ proposed in \cite{zhang2018unreasonable} and choose VGG \cite{simonyan2014very} as the backbone of LPIPS. The perceptual loss significantly improves the details of rendered results, and the structure regularization it brings helps stabilize the training process as well. The total loss is defined as:
\begin{equation}
  \mathcal{L} = \mathcal{L}_{C} + \lambda_{\textrm{lpips}} \mathcal{L}_{\textrm{lpips}}
  \label{eq:total-loss}
\end{equation}

%-------------------------------------------------------------------------
\subsection{Implementation Details}
We implement our network with PyTorch~\cite{paszke2019pytorch}, conduct mesh rasterization using PyTorch3D~\cite{ravi2020accelerating} and keep the differential Gaussian rasterization presented by 3D-GS~\cite{kerbl3Dgaussians}. For FLAME tracking, we use the analysis-by-synthesis-based face tracker from MICA~\cite{MICA:ECCV2022} further modified in INSTA~\cite{zielonka2023instant}. And the expression code $\psi$ is the concatenation of tracked expression coefficients, eyes pose, jaw pose, and eyelids coefficients.

\noindent
\textbf{Gaussian initialization and deformation.}
% We add extra $30$ faces to close the mouth cavity of FLAME mesh and adjust UV parameterization with the help of Blender.
We set the UV map resolution to $128$, sample every UV pixel with correspondence to the head region, including the neck, and the total Gaussian number is $13453$. We set the depth of offset MLP $D=5$ and the dimension of hidden layer $W=256$.

\noindent
\textbf{Optimization.}
Parameters required to be optimized include attributes of 3D Gaussians except for position and parameters of the offset network. We train our models using an Adam optimizer~\cite{kingma2014adam} with $\beta = (0.9, 0.999)$. The learning rate of Gaussians' parameters is the same as the official implementation, while the learning rate of the offset network is $\eta = 1e-4$. We choose $\lambda_{\textrm{mouth}} = 40$ and we set $\lambda_{\textrm{lpips}}$ to $0$ in the first $15000$ training steps and $0.05$ later. For each epoch, we randomly sample 2000 frames from the training dataset for training.

\section{Experiments}
\label{sec:experiments}

\begin{figure}
  \centering
  \includegraphics[width=1.0\linewidth]{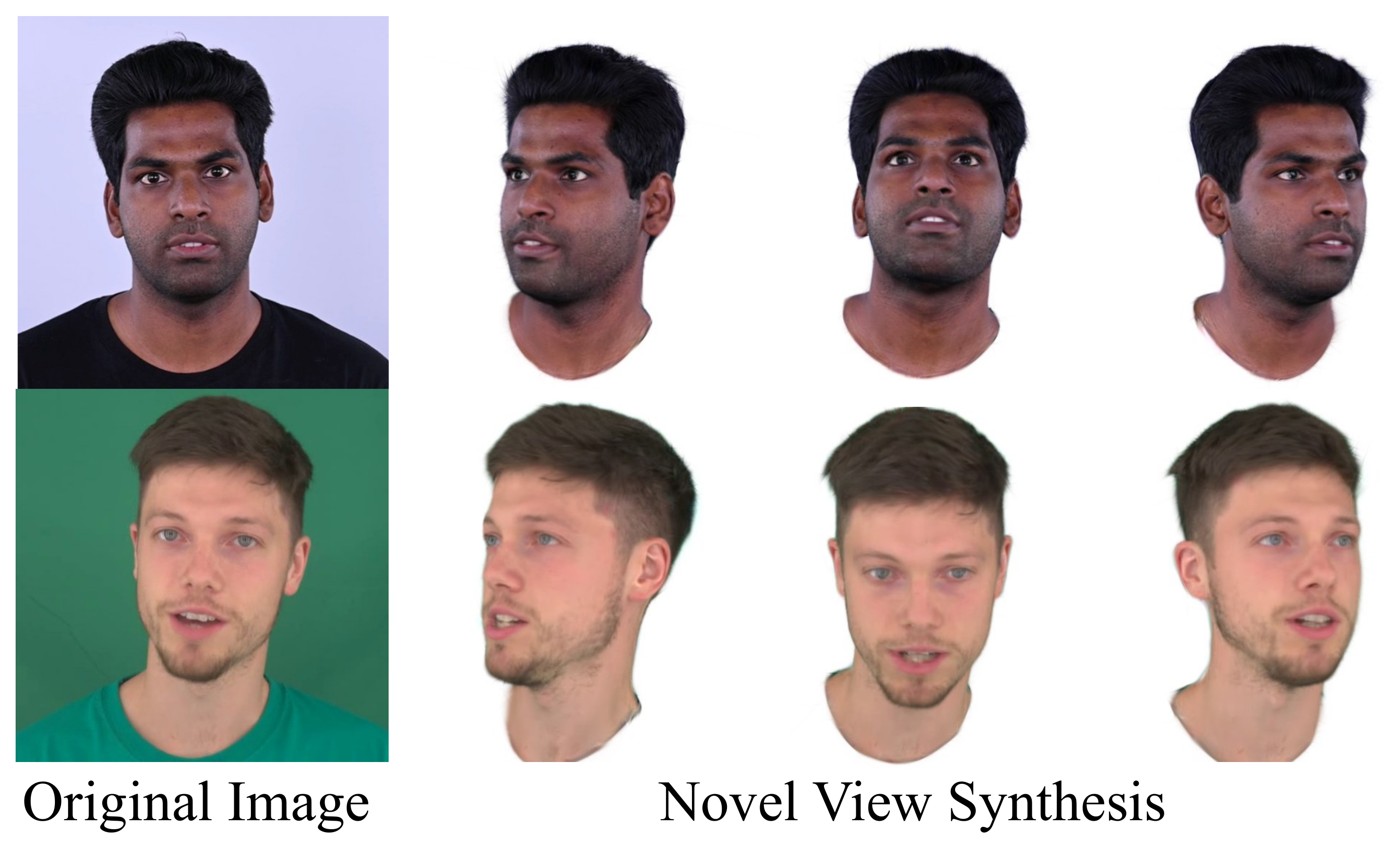}
  \caption{Our model builds on a non-neural Gaussian field and shows excellent 3D consistency.}
  \vspace*{-3mm}
  \label{fig:nvs}
\end{figure}

\begin{figure}
  \centering
  \includegraphics[width=1.0\linewidth]{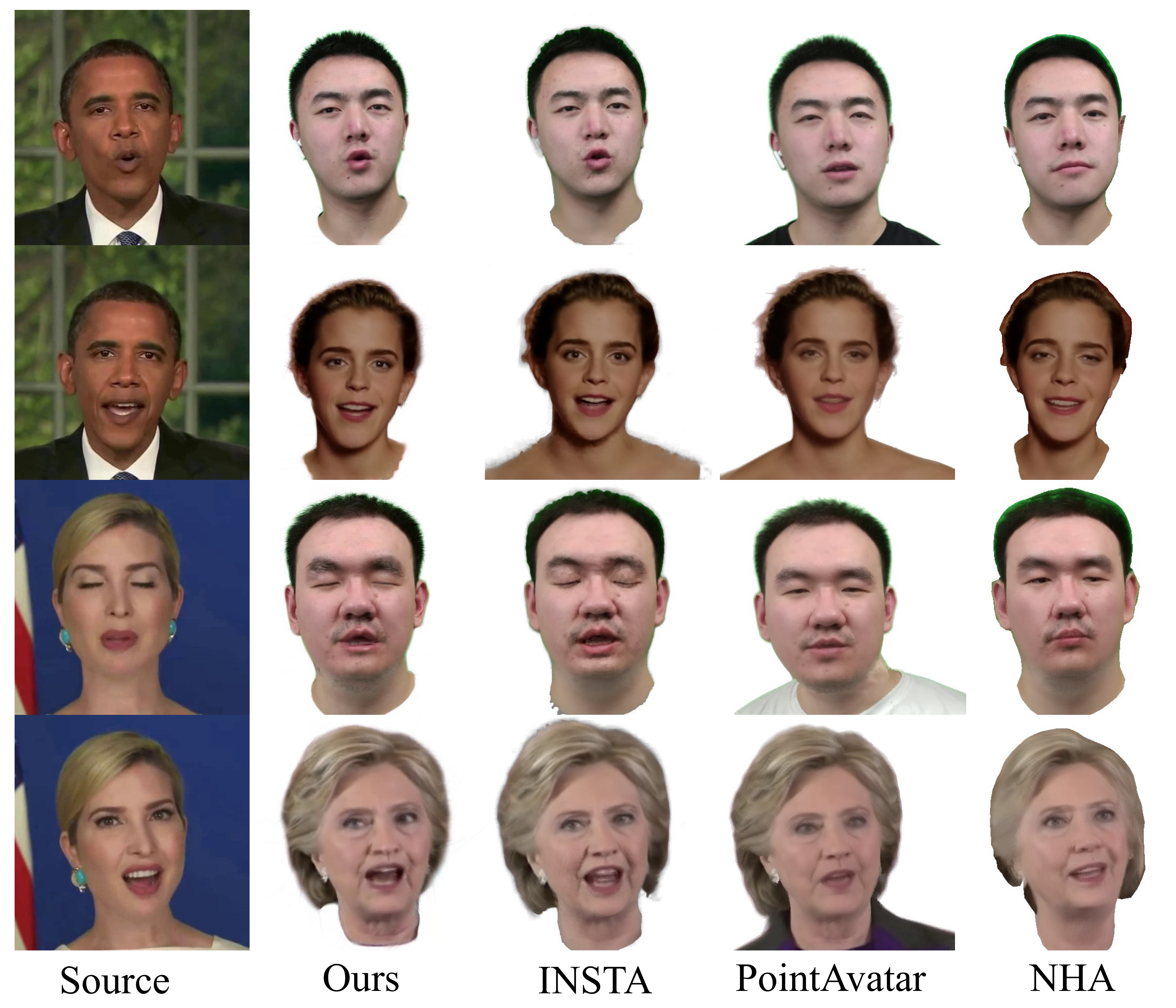}
  \caption{Qualitative results of ours and three other methods on facial reenactment task. Our method preserves personalized facial details in hair, eyes, and interior mouth regions and synthesizes more natural results.}
  \vspace*{-3mm}
  \label{fig:driving}
\end{figure}

\begin{table}
  \centering
  \begin{tabular}{@{}lcccc@{}}
    \toprule
    Metrics & NHA & PointAvatar & INSTA & Ours \\
    \midrule
    MSE($10^{-3}$)$\downarrow$ & 1.49 & 2.47 & 0.95 & \textbf{0.66} \\
    L1($10^{-2}$)$\downarrow$ & 0.99 & 1.52 & 0.89 & \textbf{0.83} \\
    PSNR$\uparrow$ & 28.80 & 27.03 & 30.54 & \textbf{32.33} \\
    SSIM($10^{-1}$)$\uparrow$ & 9.31 & 9.00 & 9.40 & \textbf{9.42} \\
    LPIPS($10^{-2}$)$\downarrow$ & 4.01 & 5.89 & 3.76 & \textbf{3.23} \\
    \bottomrule
  \end{tabular}
  \caption{Quantitative comparisons with state-of-the-art head avatar reconstruction methods on public data released by previous works. Our method outperforms others both in pixel-wise error metrics and perceptual quality.}
  % \vspace*{-3mm}
  \label{tab:psnr}
\end{table}
\begin{figure*}
  \centering
  \includegraphics[width=1.0\linewidth]{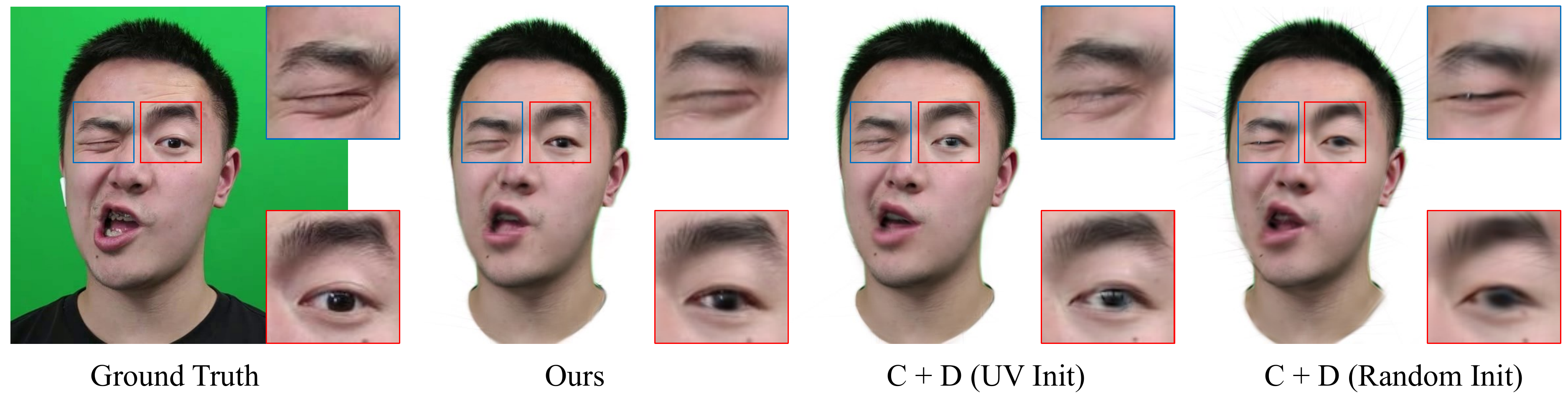}
  \caption{Comparison with “canonical + deformation” strategy. This strategy could get better results with the help of our uniform UV sampling but still fails to capture subtle expression details as well as ours.}
  \label{fig:cano}
\end{figure*}

%-------------------------------------------------------------------------
\subsection{Dataset}
% Given a monocular video, we are able to recover the corresponding 3d head avatar which is animatable as well.
To prove the robustness and fidelity of our methods, We mainly use the data released by previous works~\cite{gao2022reconstructing,zielonka2023instant,thies2020neural,grassal2022neural}, and we appreciate a lot for their sharing.
% To increase the diversity of data, we also collect some videos from YouTube for experiments. 
All videos are cropped, sub-sampled to $25$ FPS, and resized to $512^2$ resolution in advance.
The length of the processed video is between 1 and 3 minutes, and we use the last 500 frames as the testing dataset.
We use RVM~\cite{lin2022robust} for foreground segmentation and an off-the-shelf face parsing framework~\cite{yu2021bisenet} for mouth region parsing.

%-------------------------------------------------------------------------
\subsection{Comparison with Representative Methods}
We compare our method with three representative works, including (1) neural head avatar (NHA)~\cite{grassal2022neural}, typical work of explicit mesh-based methods; (2) PointAvatar~\cite{zheng2023pointavatar}, modeling the head geometry with particle-based representation (\ie point clouds) similar to us; and (3) INSTA~\cite{zielonka2023instant}, representative of efficient implicit head representation which creates a surface-embedded dynamic neural radiance field based on neural graphics primitives.
Note that for PointAvatar, the full training requires 80GB A100 GPU, but we train it on 32GB V100 and use fewer points and earlier checkpoints exactly following the author's suggestions.
All other experiments were done on 24GB Nvidia RTX 3090.
% We didn't compare with NeRFBlendshape~\cite{gao2022reconstructing} and AvatarMAV~\cite{xu2023avatarmav} because they don't include neck region. The two methods and INSTA all emphasize training acceleration, while our method is not only on par with them in training speed, but also far surpasses them in rendering speed.
NeRFBlendshape~\cite{gao2022reconstructing}, AvatarMAV~\cite{xu2023avatarmav}, and INSTA all emphasize training acceleration. We choose INSTA for comparison as it provides tracking code, models neck region, and is almost the latest work among them. FlashAvatar is not only on par with them in training efficiency but also far surpasses them in rendering speed.

\cref{fig:comparison} depicts the qualitative comparison between our model and the above methods.
% and it can be observed that our method is superior to others. 
As we can see, the representation ability of NHA is restricted by the explicit mesh domain, and it may generate undesired geometric artifacts.
INSTA uses neural graphics primitives embedded around the FLAME surface and thus cannot well model accessories like eyeglasses and earphones. Also, it tends to generate smooth results and ignore thin structures, especially in the hair region.
As for PointAvatar, the stack of points could recover glasses and earphones, but it still fails to model subtle expressions and clear teeth even with huge memory consumption. 
In contrast, our method produces photo-realistic images most consistent with the ground truth. We recover almost all fine facial details, thin structures, and subtle expressions with 3D Gaussians in 10K level.

\cref{tab:psnr} shows the quantitative comparison between our model and other methods. We compute the average errors of tested videos. The metrics include Mean Squared Error (MSE), L1 distance, PSNR, SSIM, and LPIPS \cite{zhang2018unreasonable}. 

As both mesh dynamics and later Gaussian deformation condition on tracked expression code disentangled from identity space, we could conduct facial reenactment task at super-fast rendering speed with no difficulty. We show the result of compared methods and ours in \cref{fig:driving}. Also, the basic representation of 3D head avatars in our method is pure non-neural 3D Gaussians, so we can freely adjust the global camera pose to generate target results with any desired rendering view (see \cref{fig:nvs}).
%-------------------------------------------------------------------------
\begin{figure*}
  \centering
  \includegraphics[width=1.0\linewidth]{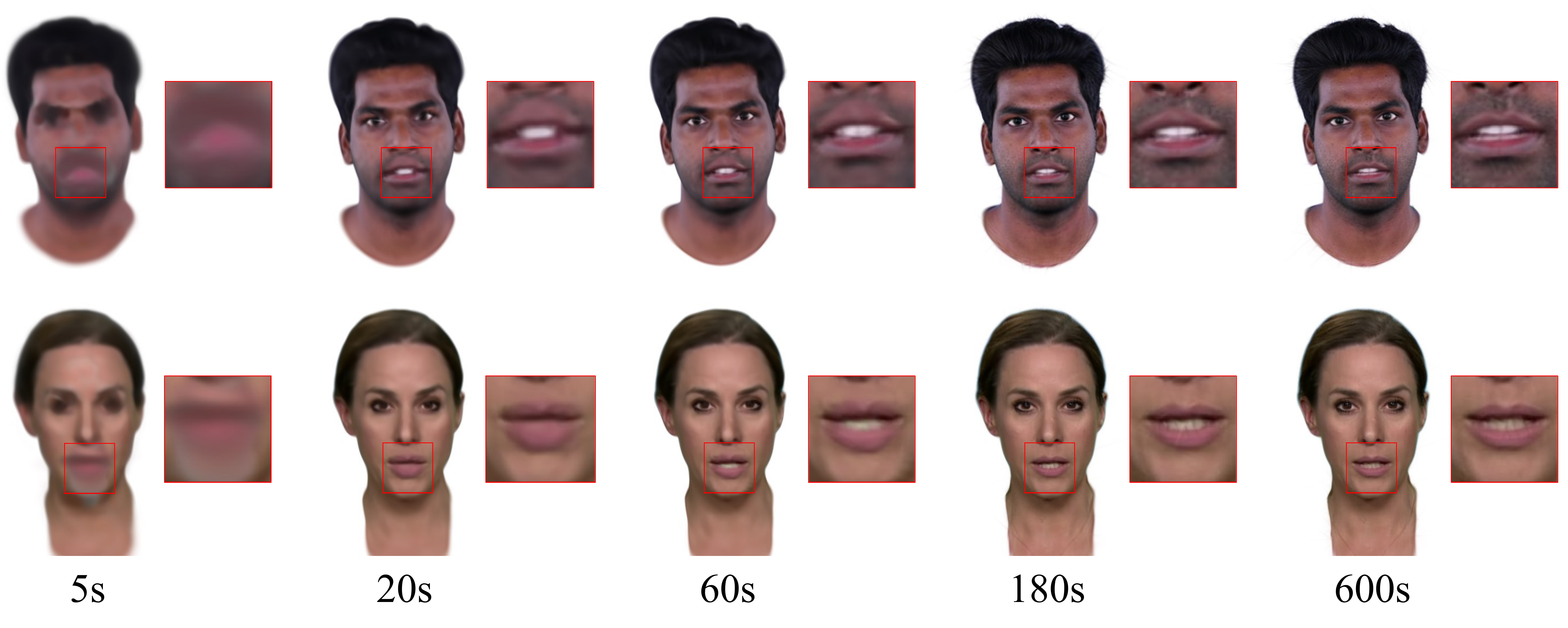}
  \caption{Besides significantly fast rendering speed at 300FPS, our training process is also efficient. High-frequency details like hair strands and teeth are fully reconstructed within a few minutes.}
  \label{fig:time}
\end{figure*}
\begin{figure}
  \centering
  \includegraphics[width=1.0\linewidth]{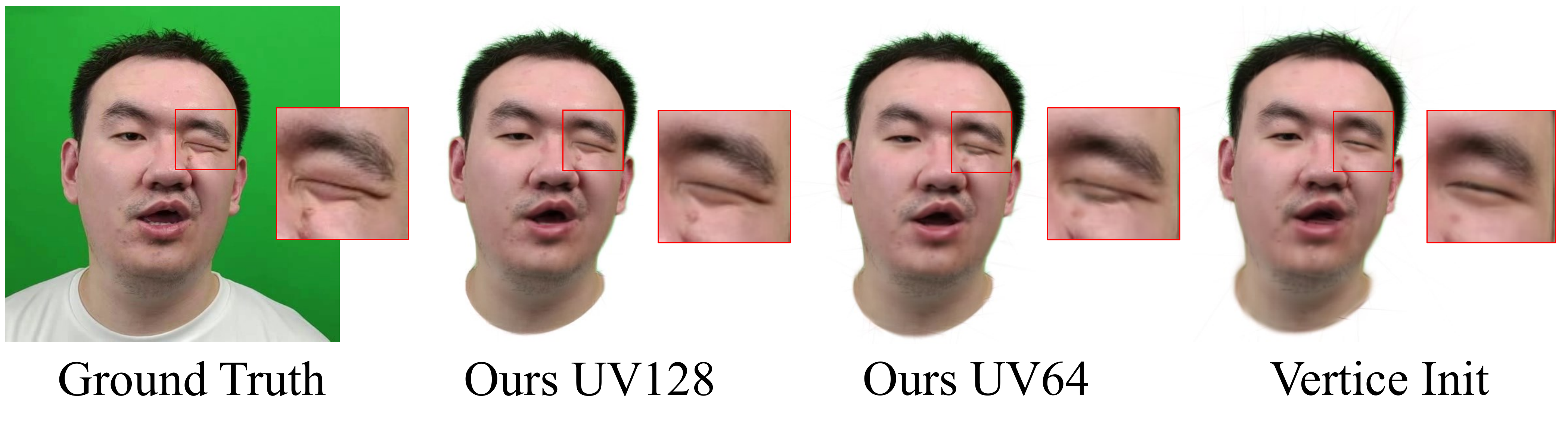}
  \caption{More uniform face initialization leads to better results than vertice initialization.}
  \vspace*{-4mm}
  \label{fig:vertice}
\end{figure}
\subsection{Comparison with C + D strategy}
While the ``canonical + deformation'' (C + D) strategy is a common way to model dynamics, it struggles to model complex expressions accurately and capture all facial details, especially when we restrict the number of Gaussians to a low level (see~\cref{fig:cano}). Following former works~\cite{yang2023deformable3dgs,wu20234dgaussians}, we randomly initialize the Gaussians in a ball (scaled by the mean size of the head), solely train the canonical 3D Gaussians during the initial 3k iterations and then jointly train Gaussians and the deformation field. However, this common strategy fails to get an acceptable head avatar with many artifacts existing, especially around the head edges. And if we introduce partial geometry priors by initializing canonical Gaussians on the mesh surface the same as ours, most artifacts disappear, but subtle expression details are still not well captured.
By comparison, we just need to model extra offset on the basis of a mesh-dependent Gaussian field.
Thus, our method can hold exaggerated expressions and preserve fine details with the help of mesh geometry guidance.
%-------------------------------------------------------------------------
\subsection{Training Efficiency}
We achieve a remarkable rendering speed over 300FPS. Meanwhile, we demonstrate that our training process is super efficient as well in~\cref{fig:time}. We are able to recover the coarse appearance of head in several seconds and reconstruct the photo-realistic avatar with fine hair strands and textures within a couple of minutes. We conduct both training and inference on a single Nvidia RTX 3090.
% which is commonly equipped.\gaoxuan{delete commonly equipped?}
%-------------------------------------------------------------------------
\subsection{Ablation Studies}
\noindent
\textbf{Gaussian Sampling Density.}
We mainly control the density of Gaussians by adjusting resolutions of the UV map, and~\cref{tab:density} shows the influence of Gaussian sampling density. While sampling more Gaussians will lead to quality improvement, it will also slow down rendering speed. We set UV resolution to 128 but also advise adjusting sampling density according to specific needs.
\begin{table}[h]
  \centering
  \begin{tabular}{@{}lcccc@{}}
    \toprule
    UV Resolution & PSNR$\uparrow$ & LPIPS($10^{-2}$)$\downarrow$ & FPS & \# GS \\
    \midrule
    64 & 30.35 & 4.47 & 394 & 3348\\
    \textbf{128} & 30.80 & 3.47 & 304 & 13453\\
    256 & 31.07 & 2.99 & 112 & 53678\\
    \bottomrule
  \end{tabular}
  \caption{Influence of Gaussian density. We set the UV Resolution to 128 in the process of comparison.}
  \label{tab:density}
\end{table}

\noindent
\textbf{Surface Embedding Methods.}
Attaching Gaussians to mesh vertices cannot converge to satisfactory results (see~\cref{fig:vertice}).
% Through our uniform UV sampling, we can get photo-realistic results even at a low Gaussian density.
Gaussian initialization in UV space is much more uniform than vertice initialization and thus we could get more photo-realistic results with fewer 3D Gaussians.

Distributing Gaussians more carefully or adaptively according to the complexity of different regions and semantic correspondence could get better results, but~\cref{tab:density} has also shown that further processing like local pruning or densification can only get slight improvement on rendering quality and speed on the base of our settings.

\begin{figure}
  \centering
  \includegraphics[width=1.0\linewidth]{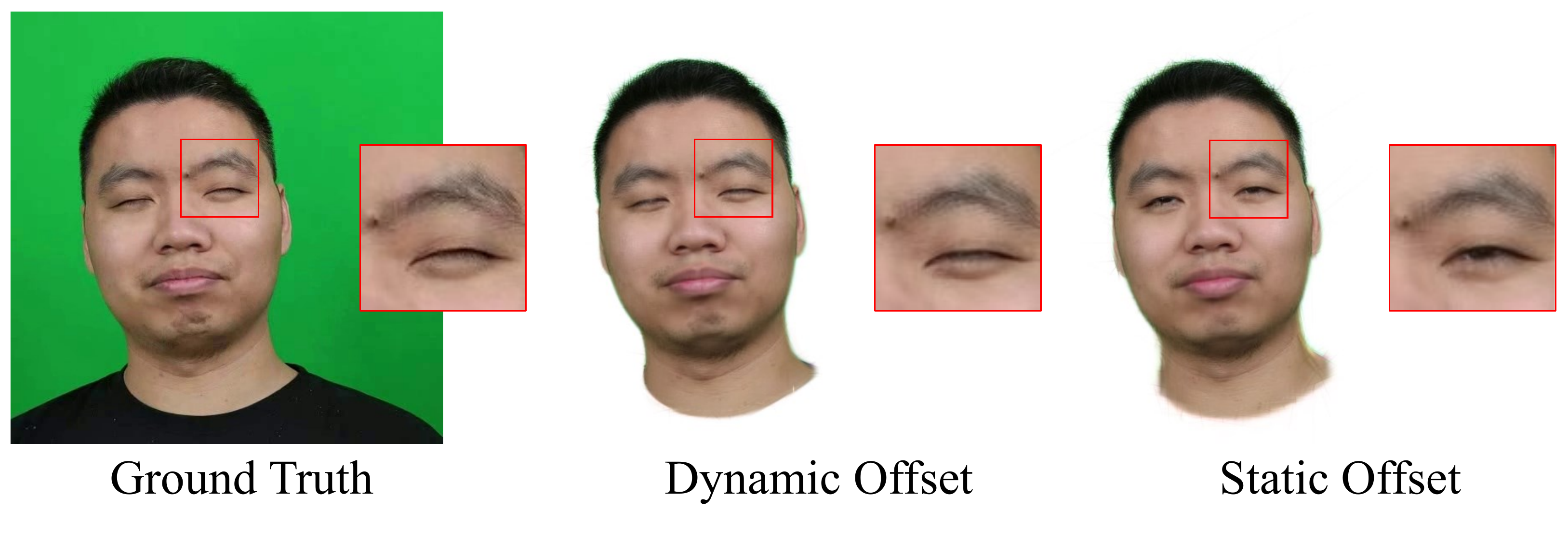}
  \caption{A dynamic offset field is of great importance to modeling fine facial expressions.}
  \vspace*{-5mm}
  \label{fig:static}
\end{figure}

\noindent
\textbf{Dynamic Offset.}
Although optimizing a static offset field could well reconstruct static areas like hair regions, it fails to well model facial alterations due to the coarse geometry of FLAME mesh and the complexity of facial expression. As shown in ~\cref{fig:static}, better visual results with higher fidelity can be obtained by learning a dynamic offset field conditional on expression code.

\section{Conclusion and Discussion}
\label{sec:conclusion}
In this paper, we have proposed FlashAvatar, which tightly combines a non-neural Gaussian-based radiance field with an explicit parametric face model and takes full advantage of their respective strengths. As a result, it can reconstruct a digital avatar from a monocular video in minutes and animate it at 300FPS while achieving photo-realistic rendering with full personalized details. Its efficiency, robustness, and representation ability have also been verified by extensive experimental results.

\noindent\textbf{Limitations and Future Work.} Our method still has several challenges that need to be addressed in future work.
While learning Gaussian offset could compensate for the inaccuracy of tracked mesh surface, our method still relies on a good surface-embedded Gaussian initialization. Therefore, large errors in tracking, especially global pose errors, may cause loss of details or image misalignment. Besides, our representation conditions on tracked expression code and thus cannot model dynamically changing hairs with heavy non-rigid deformation.

Existing works struggle to achieve real-time frame rates for high-fidelity inference, even on high-end hardware. In contrast, FlashAvatar achieves a much faster rendering speed at 300FPS on a consumer-grade GPU with SOTA rendering quality. Therefore, there will be more room for other processes in real-time tasks for multimodal digital humans, such as speech processing, text understanding, and cross-modal translation, with the help of FlashAvatar. One of our future works is to explore its potential in scenarios on mobile and mixed reality devices. We believe that our work is a solid step forward in research and practical applications of multimodal digital humans.

\noindent\textbf{Acknowledgements.} This work was supported by the National Natural Science Foundation of China (No. 62122071, No. 62272433) and the Youth Innovation Promotion Association CAS (No. 2018495).
{
    \small
    \bibliographystyle{ieeenat_fullname}
    \bibliography{main}

\begin{thebibliography}{63}
\providecommand{\natexlab}[1]{#1}
\providecommand{\url}[1]{\texttt{#1}}
\expandafter\ifx\csname urlstyle\endcsname\relax
  \providecommand{\doi}[1]{doi: #1}\else
  \providecommand{\doi}{doi: \begingroup \urlstyle{rm}\Url}\fi

\bibitem[Athar et~al.(2022)Athar, Xu, Sunkavalli, Shechtman, and Shu]{athar2022rignerf}
ShahRukh Athar, Zexiang Xu, Kalyan Sunkavalli, Eli Shechtman, and Zhixin Shu.
\newblock Rignerf: Fully controllable neural 3d portraits.
\newblock In \emph{Proceedings of the IEEE/CVF conference on Computer Vision and Pattern Recognition}, pages 20364--20373, 2022.

\bibitem[Barron et~al.(2022)Barron, Mildenhall, Verbin, Srinivasan, and Hedman]{barron2022mip}
Jonathan~T Barron, Ben Mildenhall, Dor Verbin, Pratul~P Srinivasan, and Peter Hedman.
\newblock Mip-nerf 360: Unbounded anti-aliased neural radiance fields.
\newblock In \emph{Proceedings of the IEEE/CVF Conference on Computer Vision and Pattern Recognition}, pages 5470--5479, 2022.

\bibitem[Bergman et~al.(2022)Bergman, Kellnhofer, Yifan, Chan, Lindell, and Wetzstein]{bergman2022generative}
Alexander Bergman, Petr Kellnhofer, Wang Yifan, Eric Chan, David Lindell, and Gordon Wetzstein.
\newblock Generative neural articulated radiance fields.
\newblock \emph{Advances in Neural Information Processing Systems}, 35:\penalty0 19900--19916, 2022.

\bibitem[Blanz and Vetter(1999)]{blanz1999morphable}
Volker Blanz and Thomas Vetter.
\newblock A morphable model for the synthesis of 3d faces.
\newblock In \emph{Proceedings of the 26th Annual Conference on Computer Graphics and Interactive Techniques (SIGGRAPH)}, pages 187--194, 1999.

\bibitem[Cao et~al.(2013)Cao, Weng, Zhou, Tong, and Zhou]{cao2013facewarehouse}
Chen Cao, Yanlin Weng, Shun Zhou, Yiying Tong, and Kun Zhou.
\newblock Facewarehouse: A 3d facial expression database for visual computing.
\newblock \emph{IEEE Transactions on Visualization and Computer Graphics}, 20\penalty0 (3):\penalty0 413--425, 2013.

\bibitem[Chen et~al.(2022)Chen, Xu, Geiger, Yu, and Su]{chen2022tensorf}
Anpei Chen, Zexiang Xu, Andreas Geiger, Jingyi Yu, and Hao Su.
\newblock Tensorf: Tensorial radiance fields.
\newblock In \emph{European Conference on Computer Vision}, pages 333--350. Springer, 2022.

\bibitem[Chen et~al.(2023)Chen, Wang, and Liu]{chen2023text}
Zilong Chen, Feng Wang, and Huaping Liu.
\newblock Text-to-3d using gaussian splatting.
\newblock \emph{arXiv preprint arXiv:2309.16585}, 2023.

\bibitem[Feng et~al.(2021)Feng, Feng, Black, and Bolkart]{DECA:Siggraph2021}
Yao Feng, Haiwen Feng, Michael~J. Black, and Timo Bolkart.
\newblock Learning an animatable detailed {3D} face model from in-the-wild images.
\newblock 2021.

\bibitem[Fridovich-Keil et~al.(2022)Fridovich-Keil, Yu, Tancik, Chen, Recht, and Kanazawa]{fridovich2022plenoxels}
Sara Fridovich-Keil, Alex Yu, Matthew Tancik, Qinhong Chen, Benjamin Recht, and Angjoo Kanazawa.
\newblock Plenoxels: Radiance fields without neural networks.
\newblock In \emph{Proceedings of the IEEE/CVF Conference on Computer Vision and Pattern Recognition}, pages 5501--5510, 2022.

\bibitem[Gafni et~al.(2021)Gafni, Thies, Zollhofer, and Nie{\ss}ner]{gafni2021dynamic}
Guy Gafni, Justus Thies, Michael Zollhofer, and Matthias Nie{\ss}ner.
\newblock Dynamic neural radiance fields for monocular 4d facial avatar reconstruction.
\newblock In \emph{Proceedings of the IEEE/CVF Conference on Computer Vision and Pattern Recognition}, pages 8649--8658, 2021.

\bibitem[Gao et~al.(2022)Gao, Zhong, Xiang, Hong, Guo, and Zhang]{gao2022reconstructing}
Xuan Gao, Chenglai Zhong, Jun Xiang, Yang Hong, Yudong Guo, and Juyong Zhang.
\newblock Reconstructing personalized semantic facial nerf models from monocular video.
\newblock \emph{ACM Transactions on Graphics (TOG)}, 41\penalty0 (6):\penalty0 1--12, 2022.

\bibitem[Garbin et~al.(2021)Garbin, Kowalski, Johnson, Shotton, and Valentin]{garbin2021fastnerf}
Stephan~J Garbin, Marek Kowalski, Matthew Johnson, Jamie Shotton, and Julien Valentin.
\newblock Fastnerf: High-fidelity neural rendering at 200fps.
\newblock In \emph{Proceedings of the IEEE/CVF International Conference on Computer Vision}, pages 14346--14355, 2021.

\bibitem[Grassal et~al.(2022)Grassal, Prinzler, Leistner, Rother, Nie{\ss}ner, and Thies]{grassal2022neural}
Philip-William Grassal, Malte Prinzler, Titus Leistner, Carsten Rother, Matthias Nie{\ss}ner, and Justus Thies.
\newblock Neural head avatars from monocular rgb videos.
\newblock In \emph{Proceedings of the IEEE/CVF Conference on Computer Vision and Pattern Recognition}, pages 18653--18664, 2022.

\bibitem[Guo et~al.(2021{\natexlab{a}})Guo, Cai, and Zhang]{guo20213d}
Yudong Guo, Lin Cai, and Juyong Zhang.
\newblock 3d face from {X:} learning face shape from diverse sources.
\newblock \emph{{IEEE} Trans. Image Process.}, 30:\penalty0 3815--3827, 2021{\natexlab{a}}.

\bibitem[Guo et~al.(2021{\natexlab{b}})Guo, Chen, Liang, Liu, Bao, and Zhang]{guo2021adnerf}
Yudong Guo, Keyu Chen, Sen Liang, Yongjin Liu, Hujun Bao, and Juyong Zhang.
\newblock Ad-nerf: Audio driven neural radiance fields for talking head synthesis.
\newblock In \emph{{IEEE/CVF} International Conference on Computer Vision (ICCV)}, 2021{\natexlab{b}}.

\bibitem[Hedman et~al.(2021)Hedman, Srinivasan, Mildenhall, Barron, and Debevec]{hedman2021baking}
Peter Hedman, Pratul~P Srinivasan, Ben Mildenhall, Jonathan~T Barron, and Paul Debevec.
\newblock Baking neural radiance fields for real-time view synthesis.
\newblock In \emph{Proceedings of the IEEE/CVF International Conference on Computer Vision}, pages 5875--5884, 2021.

\bibitem[Hong et~al.(2022)Hong, Peng, Xiao, Liu, and Zhang]{hong2022headnerf}
Yang Hong, Bo Peng, Haiyao Xiao, Ligang Liu, and Juyong Zhang.
\newblock Headnerf: A real-time nerf-based parametric head model.
\newblock In \emph{Proceedings of the IEEE/CVF Conference on Computer Vision and Pattern Recognition}, pages 20374--20384, 2022.

\bibitem[Huber(1992)]{huber1992robust}
Peter~J Huber.
\newblock Robust estimation of a location parameter.
\newblock In \emph{Breakthroughs in statistics: Methodology and distribution}, pages 492--518. Springer, 1992.

\bibitem[Jiang et~al.(2022)Jiang, Hong, Bao, and Zhang]{jiang2022selfrecon}
Boyi Jiang, Yang Hong, Hujun Bao, and Juyong Zhang.
\newblock Selfrecon: Self reconstruction your digital avatar from monocular video.
\newblock In \emph{{IEEE/CVF} Conference on Computer Vision and Pattern Recognition (CVPR)}, 2022.

\bibitem[Kerbl et~al.(2023)Kerbl, Kopanas, Leimk{\"u}hler, and Drettakis]{kerbl3Dgaussians}
Bernhard Kerbl, Georgios Kopanas, Thomas Leimk{\"u}hler, and George Drettakis.
\newblock 3d gaussian splatting for real-time radiance field rendering.
\newblock \emph{ACM Transactions on Graphics}, 42\penalty0 (4), 2023.

\bibitem[Khakhulin et~al.(2022)Khakhulin, Sklyarova, Lempitsky, and Zakharov]{khakhulin2022realistic}
Taras Khakhulin, Vanessa Sklyarova, Victor Lempitsky, and Egor Zakharov.
\newblock Realistic one-shot mesh-based head avatars.
\newblock In \emph{European Conference on Computer Vision}, pages 345--362. Springer, 2022.

\bibitem[Kim et~al.(2018)Kim, Garrido, Tewari, Xu, Thies, Niessner, P{\'e}rez, Richardt, Zollh{\"o}fer, and Theobalt]{kim2018deep}
Hyeongwoo Kim, Pablo Garrido, Ayush Tewari, Weipeng Xu, Justus Thies, Matthias Niessner, Patrick P{\'e}rez, Christian Richardt, Michael Zollh{\"o}fer, and Christian Theobalt.
\newblock Deep video portraits.
\newblock \emph{ACM transactions on graphics (TOG)}, 37\penalty0 (4):\penalty0 1--14, 2018.

\bibitem[Kingma and Ba(2014)]{kingma2014adam}
Diederik~P Kingma and Jimmy Ba.
\newblock Adam: A method for stochastic optimization.
\newblock \emph{arXiv preprint arXiv:1412.6980}, 2014.

\bibitem[Kopanas et~al.(2021)Kopanas, Philip, Leimk{\"u}hler, and Drettakis]{kopanas2021point}
Georgios Kopanas, Julien Philip, Thomas Leimk{\"u}hler, and George Drettakis.
\newblock Point-based neural rendering with per-view optimization.
\newblock In \emph{Computer Graphics Forum}, pages 29--43. Wiley Online Library, 2021.

\bibitem[Li et~al.(2017)Li, Bolkart, Black, Li, and Romero]{li2017learning}
Tianye Li, Timo Bolkart, Michael~J Black, Hao Li, and Javier Romero.
\newblock Learning a model of facial shape and expression from 4d scans.
\newblock \emph{ACM Trans. Graph.}, 36\penalty0 (6):\penalty0 194--1, 2017.

\bibitem[Lin et~al.(2022)Lin, Yang, Saleemi, and Sengupta]{lin2022robust}
Shanchuan Lin, Linjie Yang, Imran Saleemi, and Soumyadip Sengupta.
\newblock Robust high-resolution video matting with temporal guidance.
\newblock In \emph{Proceedings of the IEEE/CVF Winter Conference on Applications of Computer Vision}, pages 238--247, 2022.

\bibitem[Liu et~al.(2020)Liu, Gu, Zaw~Lin, Chua, and Theobalt]{liu2020neural}
Lingjie Liu, Jiatao Gu, Kyaw Zaw~Lin, Tat-Seng Chua, and Christian Theobalt.
\newblock Neural sparse voxel fields.
\newblock \emph{Advances in Neural Information Processing Systems}, 33:\penalty0 15651--15663, 2020.

\bibitem[Lombardi et~al.(2021)Lombardi, Simon, Schwartz, Zollhoefer, Sheikh, and Saragih]{Lombardi21}
Stephen Lombardi, Tomas Simon, Gabriel Schwartz, Michael Zollhoefer, Yaser Sheikh, and Jason Saragih.
\newblock Mixture of volumetric primitives for efficient neural rendering.
\newblock \emph{ACM Trans. Graph.}, 40\penalty0 (4), 2021.

\bibitem[Luiten et~al.(2024)Luiten, Kopanas, Leibe, and Ramanan]{luiten2023dynamic}
Jonathon Luiten, Georgios Kopanas, Bastian Leibe, and Deva Ramanan.
\newblock Dynamic 3d gaussians: Tracking by persistent dynamic view synthesis.
\newblock In \emph{3DV}, 2024.

\bibitem[Mescheder et~al.(2019)Mescheder, Oechsle, Niemeyer, Nowozin, and Geiger]{mescheder2019occupancy}
Lars Mescheder, Michael Oechsle, Michael Niemeyer, Sebastian Nowozin, and Andreas Geiger.
\newblock Occupancy networks: Learning 3d reconstruction in function space.
\newblock In \emph{Proceedings of the IEEE/CVF conference on computer vision and pattern recognition}, pages 4460--4470, 2019.

\bibitem[Mildenhall et~al.(2020)Mildenhall, Srinivasan, Tancik, Barron, Ramamoorthi, and Ng]{mildenhall2020nerf}
Ben Mildenhall, Pratul~P. Srinivasan, Matthew Tancik, Jonathan~T. Barron, Ravi Ramamoorthi, and Ren Ng.
\newblock Nerf: Representing scenes as neural radiance fields for view synthesis.
\newblock In \emph{ECCV}, 2020.

\bibitem[M{\"u}ller et~al.(2022)M{\"u}ller, Evans, Schied, and Keller]{muller2022instant}
Thomas M{\"u}ller, Alex Evans, Christoph Schied, and Alexander Keller.
\newblock Instant neural graphics primitives with a multiresolution hash encoding.
\newblock \emph{ACM Transactions on Graphics (ToG)}, 41\penalty0 (4):\penalty0 1--15, 2022.

\bibitem[Niemeyer and Geiger(2021)]{niemeyer2021giraffe}
Michael Niemeyer and Andreas Geiger.
\newblock Giraffe: Representing scenes as compositional generative neural feature fields.
\newblock In \emph{Proceedings of the IEEE/CVF Conference on Computer Vision and Pattern Recognition}, pages 11453--11464, 2021.

\bibitem[Park et~al.(2019)Park, Florence, Straub, Newcombe, and Lovegrove]{park2019deepsdf}
Jeong~Joon Park, Peter Florence, Julian Straub, Richard Newcombe, and Steven Lovegrove.
\newblock Deepsdf: Learning continuous signed distance functions for shape representation.
\newblock In \emph{Proceedings of the IEEE/CVF conference on computer vision and pattern recognition}, pages 165--174, 2019.

\bibitem[Paszke et~al.(2019)Paszke, Gross, Massa, Lerer, Bradbury, Chanan, Killeen, Lin, Gimelshein, Antiga, et~al.]{paszke2019pytorch}
Adam Paszke, Sam Gross, Francisco Massa, Adam Lerer, James Bradbury, Gregory Chanan, Trevor Killeen, Zeming Lin, Natalia Gimelshein, Luca Antiga, et~al.
\newblock Pytorch: An imperative style, high-performance deep learning library.
\newblock \emph{Advances in neural information processing systems}, 32, 2019.

\bibitem[Paysan et~al.(2009)Paysan, Knothe, Amberg, Romdhani, and Vetter]{paysan20093d}
Pascal Paysan, Reinhard Knothe, Brian Amberg, Sami Romdhani, and Thomas Vetter.
\newblock A 3d face model for pose and illumination invariant face recognition.
\newblock In \emph{2009 sixth IEEE international conference on advanced video and signal based surveillance}, pages 296--301. Ieee, 2009.

\bibitem[Poole et~al.(2022)Poole, Jain, Barron, and Mildenhall]{poole2022dreamfusion}
Ben Poole, Ajay Jain, Jonathan~T Barron, and Ben Mildenhall.
\newblock Dreamfusion: Text-to-3d using 2d diffusion.
\newblock In \emph{The Eleventh International Conference on Learning Representations}, 2022.

\bibitem[Ranjan et~al.(2018)Ranjan, Bolkart, Sanyal, and Black]{ranjan2018generating}
Anurag Ranjan, Timo Bolkart, Soubhik Sanyal, and Michael~J Black.
\newblock Generating 3d faces using convolutional mesh autoencoders.
\newblock In \emph{Proceedings of the European Conference on Computer Vision (ECCV)}, pages 704--720, 2018.

\bibitem[Ravi et~al.(2020)Ravi, Reizenstein, Novotny, Gordon, Lo, Johnson, and Gkioxari]{ravi2020accelerating}
Nikhila Ravi, Jeremy Reizenstein, David Novotny, Taylor Gordon, Wan-Yen Lo, Justin Johnson, and Georgia Gkioxari.
\newblock Accelerating 3d deep learning with pytorch3d.
\newblock \emph{arXiv preprint arXiv:2007.08501}, 2020.

\bibitem[Simonyan and Zisserman(2014)]{simonyan2014very}
Karen Simonyan and Andrew Zisserman.
\newblock Very deep convolutional networks for large-scale image recognition.
\newblock \emph{arXiv preprint arXiv:1409.1556}, 2014.

\bibitem[Sun et~al.(2022)Sun, Sun, and Chen]{sun2022direct}
Cheng Sun, Min Sun, and Hwann-Tzong Chen.
\newblock Direct voxel grid optimization: Super-fast convergence for radiance fields reconstruction.
\newblock In \emph{Proceedings of the IEEE/CVF Conference on Computer Vision and Pattern Recognition}, pages 5459--5469, 2022.

\bibitem[Tang et~al.(2023)Tang, Ren, Zhou, Liu, and Zeng]{tang2023dreamgaussian}
Jiaxiang Tang, Jiawei Ren, Hang Zhou, Ziwei Liu, and Gang Zeng.
\newblock Dreamgaussian: Generative gaussian splatting for efficient 3d content creation.
\newblock \emph{arXiv preprint arXiv:2309.16653}, 2023.

\bibitem[Thies et~al.(2019)Thies, Zollh{\"o}fer, and Nie{\ss}ner]{thies2019deferred}
Justus Thies, Michael Zollh{\"o}fer, and Matthias Nie{\ss}ner.
\newblock Deferred neural rendering: Image synthesis using neural textures.
\newblock \emph{Acm Transactions on Graphics (TOG)}, 38\penalty0 (4):\penalty0 1--12, 2019.

\bibitem[Thies et~al.(2020)Thies, Elgharib, Tewari, Theobalt, and Nie{\ss}ner]{thies2020neural}
Justus Thies, Mohamed Elgharib, Ayush Tewari, Christian Theobalt, and Matthias Nie{\ss}ner.
\newblock Neural voice puppetry: Audio-driven facial reenactment.
\newblock In \emph{Computer Vision--ECCV 2020: 16th European Conference, Glasgow, UK, August 23--28, 2020, Proceedings, Part XVI 16}, pages 716--731. Springer, 2020.

\bibitem[Tran and Liu(2018)]{tran2018nonlinear}
Luan Tran and Xiaoming Liu.
\newblock Nonlinear 3d face morphable model.
\newblock In \emph{Proceedings of the IEEE conference on computer vision and pattern recognition}, pages 7346--7355, 2018.

\bibitem[Vlasic et~al.(2006)Vlasic, Brand, Pfister, and Popovic]{vlasic2006face}
Daniel Vlasic, Matthew Brand, Hanspeter Pfister, and Jovan Popovic.
\newblock Face transfer with multilinear models.
\newblock In \emph{ACM SIGGRAPH 2006 Courses}, pages 24--es. 2006.

\bibitem[Wang et~al.(2021)Wang, Liu, Liu, Theobalt, Komura, and Wang]{wang2021neus}
Peng Wang, Lingjie Liu, Yuan Liu, Christian Theobalt, Taku Komura, and Wenping Wang.
\newblock Neus: Learning neural implicit surfaces by volume rendering for multi-view reconstruction.
\newblock \emph{Advances in Neural Information Processing Systems}, 34:\penalty0 27171--27183, 2021.

\bibitem[Wu et~al.(2023)Wu, Yi, Fang, Xie, Zhang, Wei, Liu, Tian, and Xinggang]{wu20234dgaussians}
Guanjun Wu, Taoran Yi, Jiemin Fang, Lingxi Xie, Xiaopeng Zhang, Wei Wei, Wenyu Liu, Qi Tian, and Wang Xinggang.
\newblock 4d gaussian splatting for real-time dynamic scene rendering.
\newblock \emph{arXiv preprint arXiv:2310.08528}, 2023.

\bibitem[Xu et~al.(2023)Xu, Wang, Zhao, Zhang, and Liu]{xu2023avatarmav}
Yuelang Xu, Lizhen Wang, Xiaochen Zhao, Hongwen Zhang, and Yebin Liu.
\newblock Avatarmav: Fast 3d head avatar reconstruction using motion-aware neural voxels.
\newblock In \emph{ACM SIGGRAPH 2023 Conference Proceedings}, pages 1--10, 2023.

\bibitem[Yang et~al.(2020)Yang, Zhu, Wang, Huang, Shen, Yang, and Cao]{yang2020facescape}
Haotian Yang, Hao Zhu, Yanru Wang, Mingkai Huang, Qiu Shen, Ruigang Yang, and Xun Cao.
\newblock Facescape: a large-scale high quality 3d face dataset and detailed riggable 3d face prediction.
\newblock In \emph{Proceedings of the ieee/cvf conference on computer vision and pattern recognition}, pages 601--610, 2020.

\bibitem[Yang et~al.(2023)Yang, Gao, Zhou, Jiao, Zhang, and Jin]{yang2023deformable3dgs}
Ziyi Yang, Xinyu Gao, Wen Zhou, Shaohui Jiao, Yuqing Zhang, and Xiaogang Jin.
\newblock Deformable 3d gaussians for high-fidelity monocular dynamic scene reconstruction.
\newblock \emph{arXiv preprint arXiv:2309.13101}, 2023.

\bibitem[Yi et~al.(2023)Yi, Fang, Wu, Xie, Zhang, Liu, Tian, and Wang]{GaussianDreamer}
Taoran Yi, Jiemin Fang, Guanjun Wu, Lingxi Xie, Xiaopeng Zhang, Wenyu Liu, Qi Tian, and Xinggang Wang.
\newblock Gaussiandreamer: Fast generation from text to 3d gaussian splatting with point cloud priors.
\newblock \emph{arxiv:2310.08529}, 2023.

\bibitem[Yifan et~al.(2019)Yifan, Serena, Wu, {\"O}ztireli, and Sorkine-Hornung]{yifan2019differentiable}
Wang Yifan, Felice Serena, Shihao Wu, Cengiz {\"O}ztireli, and Olga Sorkine-Hornung.
\newblock Differentiable surface splatting for point-based geometry processing.
\newblock \emph{ACM Transactions on Graphics (TOG)}, 38\penalty0 (6):\penalty0 1--14, 2019.

\bibitem[Yu et~al.(2021)Yu, Gao, Wang, Yu, Shen, and Sang]{yu2021bisenet}
Changqian Yu, Changxin Gao, Jingbo Wang, Gang Yu, Chunhua Shen, and Nong Sang.
\newblock Bisenet v2: Bilateral network with guided aggregation for real-time semantic segmentation.
\newblock \emph{International Journal of Computer Vision}, 129:\penalty0 3051--3068, 2021.

\bibitem[Zhang et~al.(2020)Zhang, Riegler, Snavely, and Koltun]{kaizhang2020}
Kai Zhang, Gernot Riegler, Noah Snavely, and Vladlen Koltun.
\newblock Nerf++: Analyzing and improving neural radiance fields.
\newblock \emph{arXiv:2010.07492}, 2020.

\bibitem[Zhang et~al.(2023)Zhang, Zhao, Cong, Zhang, Gu, Gao, Zheng, Yang, Xu, and Yu]{zhang2023hack}
Longwen Zhang, Zijun Zhao, Xinzhou Cong, Qixuan Zhang, Shuqi Gu, Yuchong Gao, Rui Zheng, Wei Yang, Lan Xu, and Jingyi Yu.
\newblock Hack: Learning a parametric head and neck model for high-fidelity animation.
\newblock \emph{ACM Transactions on Graphics (TOG)}, 42\penalty0 (4):\penalty0 1--20, 2023.

\bibitem[Zhang et~al.(2018)Zhang, Isola, Efros, Shechtman, and Wang]{zhang2018unreasonable}
Richard Zhang, Phillip Isola, Alexei~A Efros, Eli Shechtman, and Oliver Wang.
\newblock The unreasonable effectiveness of deep features as a perceptual metric.
\newblock In \emph{Proceedings of the IEEE conference on computer vision and pattern recognition}, pages 586--595, 2018.

\bibitem[Zheng et~al.(2022{\natexlab{a}})Zheng, Yang, Huang, and Chen]{zheng2022imface}
Mingwu Zheng, Hongyu Yang, Di Huang, and Liming Chen.
\newblock Imface: A nonlinear 3d morphable face model with implicit neural representations.
\newblock In \emph{Proceedings of the IEEE/CVF Conference on Computer Vision and Pattern Recognition}, pages 20343--20352, 2022{\natexlab{a}}.

\bibitem[Zheng et~al.(2022{\natexlab{b}})Zheng, Abrevaya, B{\"u}hler, Chen, Black, and Hilliges]{zheng2022avatar}
Yufeng Zheng, Victoria~Fern{\'a}ndez Abrevaya, Marcel~C B{\"u}hler, Xu Chen, Michael~J Black, and Otmar Hilliges.
\newblock Im avatar: Implicit morphable head avatars from videos.
\newblock In \emph{Proceedings of the IEEE/CVF Conference on Computer Vision and Pattern Recognition}, pages 13545--13555, 2022{\natexlab{b}}.

\bibitem[Zheng et~al.(2023)Zheng, Yifan, Wetzstein, Black, and Hilliges]{zheng2023pointavatar}
Yufeng Zheng, Wang Yifan, Gordon Wetzstein, Michael~J Black, and Otmar Hilliges.
\newblock Pointavatar: Deformable point-based head avatars from videos.
\newblock In \emph{Proceedings of the IEEE/CVF Conference on Computer Vision and Pattern Recognition}, pages 21057--21067, 2023.

\bibitem[Zielonka et~al.(2022)Zielonka, Bolkart, and Thies]{MICA:ECCV2022}
Wojciech Zielonka, Timo Bolkart, and Justus Thies.
\newblock Towards metrical reconstruction of human faces.
\newblock In \emph{European Conference on Computer Vision (ECCV)}. Springer International Publishing, 2022.

\bibitem[Zielonka et~al.(2023)Zielonka, Bolkart, and Thies]{zielonka2023instant}
Wojciech Zielonka, Timo Bolkart, and Justus Thies.
\newblock Instant volumetric head avatars.
\newblock In \emph{Proceedings of the IEEE/CVF Conference on Computer Vision and Pattern Recognition}, pages 4574--4584, 2023.

\bibitem[Zwicker et~al.(2001)Zwicker, Pfister, Van~Baar, and Gross]{zwicker2001ewa}
Matthias Zwicker, Hanspeter Pfister, Jeroen Van~Baar, and Markus Gross.
\newblock Ewa volume splatting.
\newblock In \emph{Proceedings Visualization, 2001. VIS'01.}, pages 29--538. IEEE, 2001.

\end{thebibliography}
}

\clearpage
\setcounter{page}{1}
\maketitlesupplementary
\appendix

% \section{Rationale}
% \label{sec:rationale}
% % 
% Having the supplementary compiled together with the main paper means that:
% % 
% \begin{itemize}
% \item The supplementary can back-reference sections of the main paper, for example, we can refer to \cref{sec:intro};
% \item The main paper can forward reference sub-sections within the supplementary explicitly (e.g. referring to a particular experiment); 
% \item When submitted to arXiv, the supplementary will already included at the end of the paper.
% \end{itemize}
% % 
% To split the supplementary pages from the main paper, you can use \href{https://support.apple.com/en-ca/guide/preview/prvw11793/mac#:~:text=Delete%20a%20page%20from%20a,or%20choose%20Edit%20%3E%20Delete).}{Preview (on macOS)}, \href{https://www.adobe.com/acrobat/how-to/delete-pages-from-pdf.html#:~:text=Choose%20%E2%80%9CTools%E2%80%9D%20%3E%20%E2%80%9COrganize,or%20pages%20from%20the%20file.}{Adobe Acrobat} (on all OSs), as well as \href{https://superuser.com/questions/517986/is-it-possible-to-delete-some-pages-of-a-pdf-document}{command line tools}.

\section{Additional Ablations and Results}
\label{sec:AR}

\subsection{Additional Ablations}
\begin{figure}[h]
  \centering
  \includegraphics[width=1.0\linewidth]{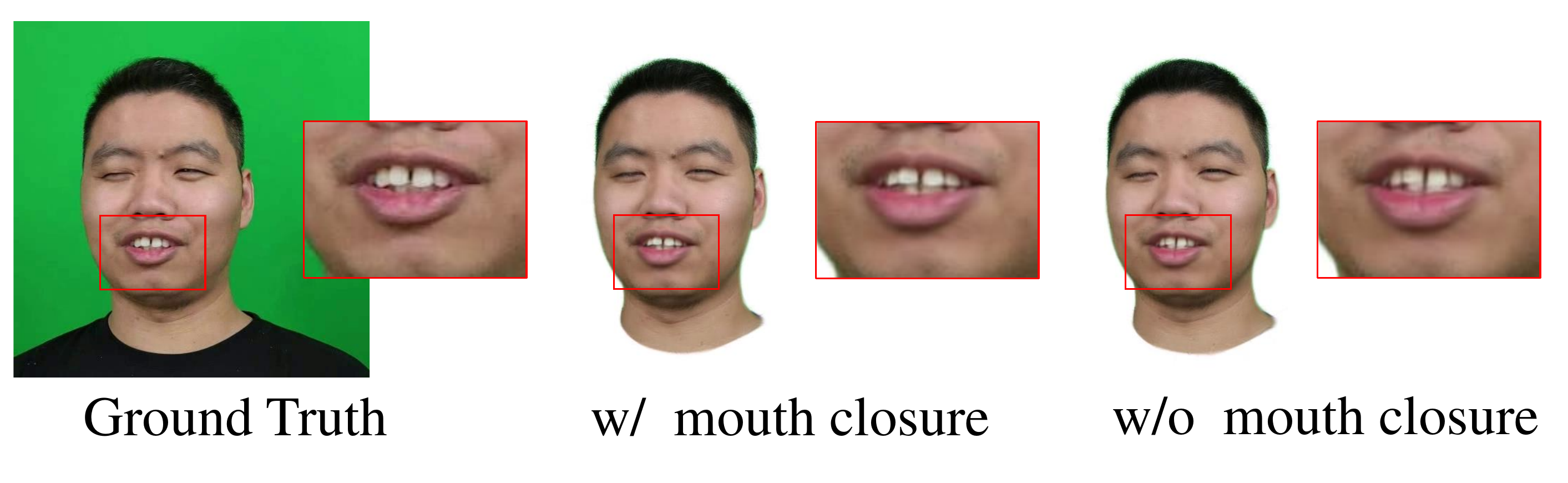}
  \caption{Closing the mouth cavity of FLAME mesh with additional faces is useful for modeling the interior mouth like teeth.}
  \label{fig:closure}
\end{figure}

\textbf{Mouth closure.}
Since the original FLAME mesh does not model the interior mouth, we add additional faces to close the mouth cavity and find it helpful in modeling the interior mouth. As seen in~\cref{fig:closure}, if we merely rely on Gaussians in nearby areas like the lips to model the interior mouth, the upper and lower teeth tend to stick together, which leads to blurry results, especially for challenging cases.

\begin{figure}[h]
  \centering
  \includegraphics[width=1.0\linewidth]{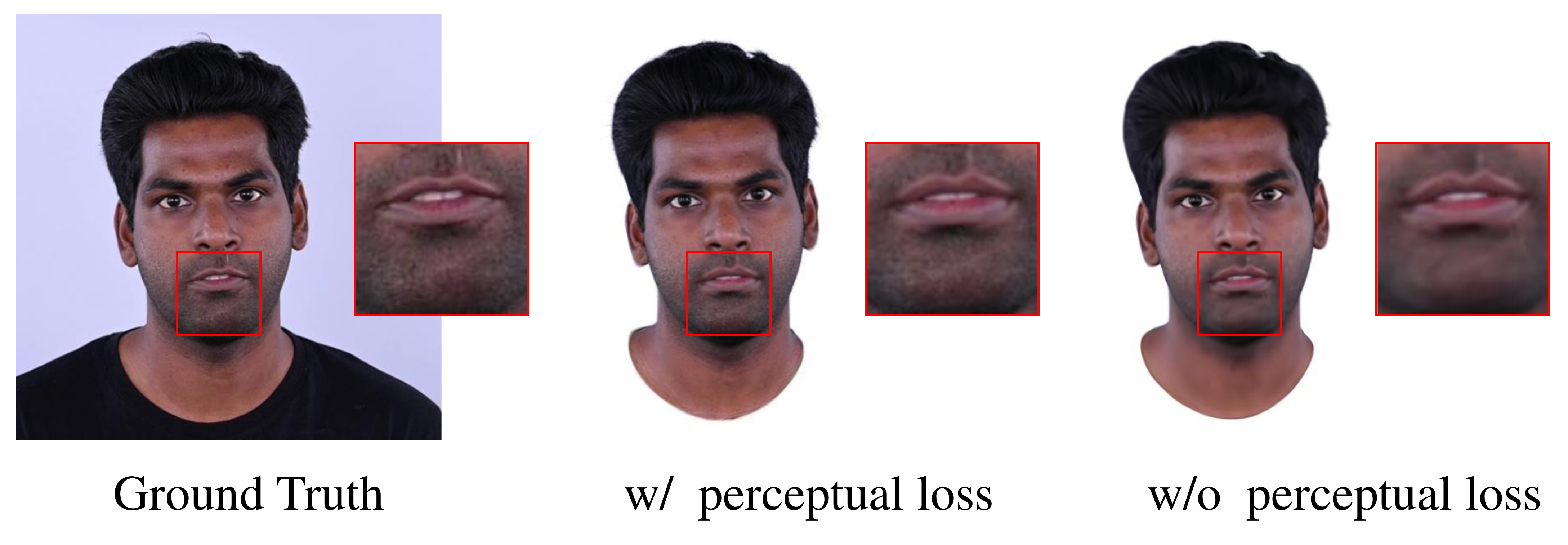}
  \caption{The perceptual loss helps maintain fine-detailed facial attributes of the head avatar.}
  \label{fig:perceptual}
\end{figure}

\noindent
\textbf{Perceptual loss.}
Besides the pixel-based loss, we adopt the perceptual loss as well. \cref{fig:perceptual} shows the comparison between the results with/without perceptual loss supervision. As we can see, the perceptual loss helps maintain personalized facial attributes and greatly boosts photo-realism.

\subsection{Additional Results}
\begin{figure}[h]
  \centering
  \includegraphics[width=1.0\linewidth]{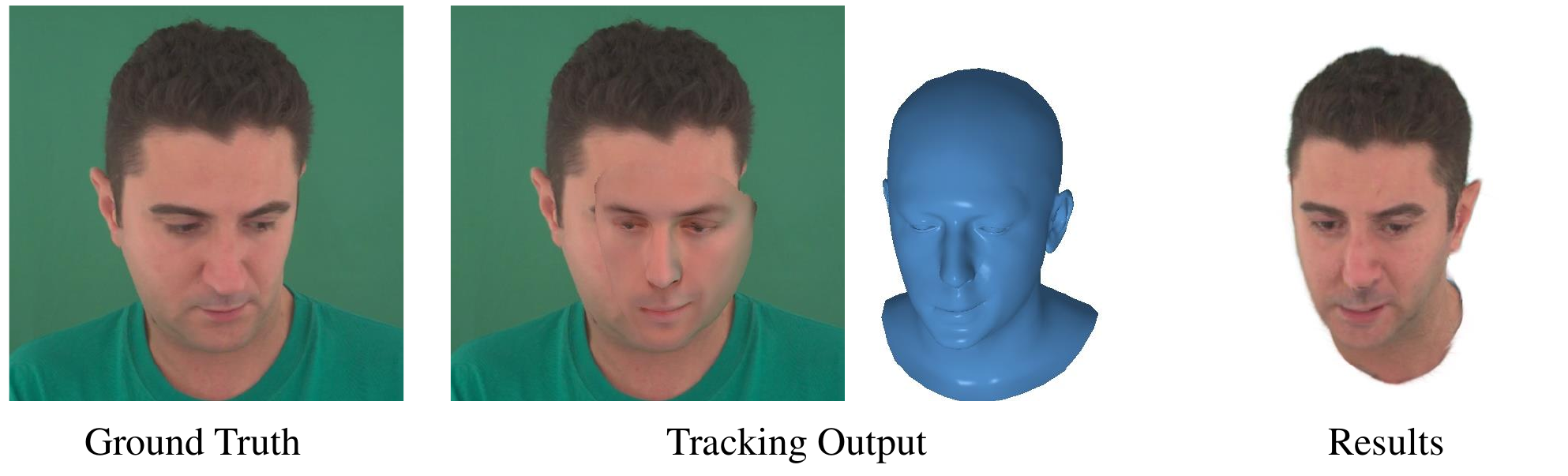}
  \caption{Large errors in tracking lead to wrong results.}
  \label{fig:error}
\end{figure}

\textbf{Limitation.}
Our method still relies on a good surface-embedded Gaussian initialization and cannot handle large errors in tracking (see~\cref{fig:error}).

%-------------------------------------------------------------------------
\section{Implementation Details}
\label{sec:ID}

\subsection{Network Architecture}
\begin{figure}[h]
  \centering
  \includegraphics[width=1.0\linewidth]{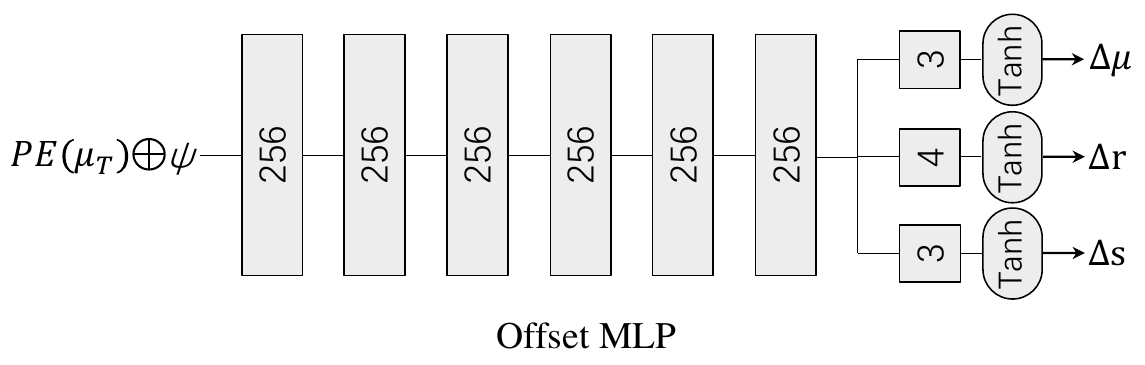}
  \caption{Network architecture of the offset MLP. Except for the last layer, each linear layer is followed by the ReLU activation.}
  \label{fig:network}
\end{figure}

We show the architecture of the offset network in~\cref{fig:network}.

\subsection{FLAME Masks}
\begin{figure}
  \centering
  \includegraphics[width=0.4\linewidth]{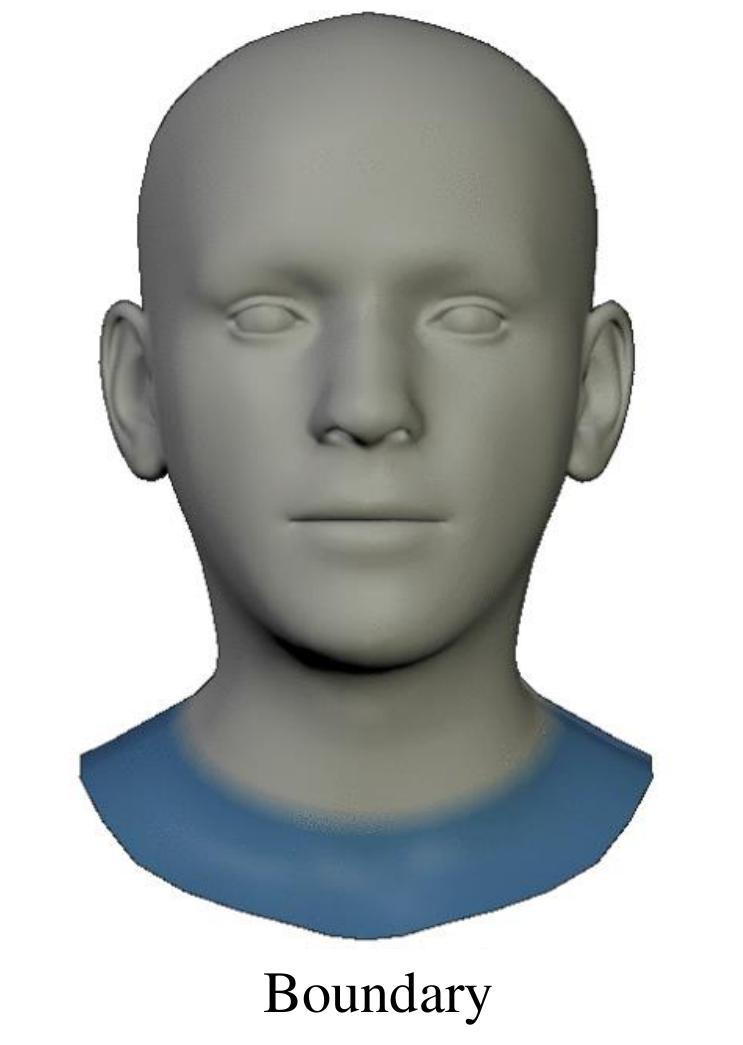}
  \caption{The blue region corresponds to the boundary of the FLAME mesh, which is excluded when sampling Gaussians.}
  \label{fig:flamemask}
\end{figure}

As we only model head regions with neck, we sample Gaussians in the corresponding areas, and we conduct this by adding a flame mask excluding the boundary of FLAME mesh (see~\cref{fig:flamemask}).
%-------------------------------------------------------------------------
\section{Broader Impact}
\label{sec:ethics}
Our work could reconstruct a digital avatar from a monocular video in minutes and animate it at 300FPS while achieving photo-realistic rendering with full personalized details. This takes an important step towards practical applications of multimodal digital humans, as it provides more space for other interactive tasks to enable real-time interaction. However, there is a risk of misuse, \eg the so-called DeepFakes. We strongly discourage using our work to generate fake images or videos of individuals with the intent of spreading false information or damaging their reputations. Unfortunately, we may be unable to prevent the nefarious use of our technology. Nevertheless, we believe that performing research in an open and transparent way could raise the public's awareness of nefarious uses, and our work could further enhance forgery detection capabilities.

\end{document}